\documentclass[a4paper,fleqn]{cas-dc}
\usepackage[authoryear]{natbib}
\usepackage{todonotes}
\usepackage{xcolor}
\usepackage{soul}
\usepackage{booktabs}
\usepackage{tabularx}
\usepackage{caption}
\usepackage{amsmath,amssymb,stmaryrd}   
\usepackage{graphicx}
\usepackage[export]{adjustbox}
\usepackage{comment}
\usepackage{longtable}
\usepackage{rotating} % Required for sidewaystable
\usepackage{float}
\usepackage{subfigure}
\usepackage{times}
\usepackage{lipsum}
\usepackage{latexsym}
\usepackage{booktabs}
\usepackage{rotating}
\usepackage{hyperref}
\usepackage[T1]{fontenc}
\usepackage{amsmath}
\usepackage{caption}
\usepackage{ulem}
\usepackage[utf8]{inputenc}
\usepackage{microtype}
\usepackage[most]{tcolorbox}
\usepackage{inconsolata}
\usepackage{multirow}

\soulregister\cite7
\def\tsc#1{\csdef{#1}{\textsc{\lowercase{#1}}\xspace}}
\tsc{WGM}
\tsc{QE}
\tsc{EP}
\tsc{PMS}
\tsc{BEC}
\tsc{DE}
\begin{document}
\let\WriteBookmarks\relax
\def\floatpagepagefraction{1}
\def\textpagefraction{.001}

% \shortauthors{Hu et~al.}
\title [mode = title]{LRP4RAG: Detecting Hallucinations in Retrieval-Augmented Generation via Layer-wise Relevance Propagation} 

% Second author
\author[1]{Haichuan Hu}[orcid=0009-0002-3007-488X]
\fnmark[1]
\ead{huhaichuan.hhc@alibaba-inc.com}

\affiliation[1]{organization={Alibaba Cloud},
    % addressline={Radarweg 29}, 
    city={Hangzhou, Zhejiang},
    % citysep={}, % Uncomment if no comma needed between city and postcode
    postcode={310058}, 
    % state={},
    country={China}}

\author[2]{Congqing He}[
                        % type=editor,
                        % auid=000,
                        % bioid=1,
                        % prefix=Sir,
                        % role=Researcher,
                        orcid=0000-0001-8361-8938
                        ]
\fnmark[1]
\ead{hecongqing@student.usm.my}

% Address/affiliation
\affiliation[2]{organization={School of Computer Sciences, Universiti Sains Malaysia},
    % addressline={Radarweg 29}, 
    city={Penang},
    % citysep={}, % Uncomment if no comma needed between city and postcode
    postcode={11800}, 
    % state={},
    country={Malaysia}}

% Third author
\author[3]{Xiaochen Xie}[%
   % role=Co-ordinator,
   % suffix=Jr,
   orcid=0009-0003-8863-9991
   ]
\ead{xcxie@zju.edu.cn}
\affiliation[3]{organization={Department of Service Science and Operations Management, School of Management, Zhejiang University},
    % addressline={Radarweg 29}, 
    city={Hangzhou, Zhejiang},
    % citysep={}, % Uncomment if no comma needed between city and postcode
    postcode={310058}, 
    % state={},
    country={China}}

\author[4]{Quanjun Zhang}[%
   % role=Co-ordinator,
   % suffix=Jr,
   orcid=0000-0002-2495-3805
   ]
\cormark[1]
\ead{quanjun.zhang@smail.nju.edu.cn}
\affiliation[4]{organization={State Key Laboratory for Novel Software Technology, Nanjing University},
    % addressline={Radarweg 29}, 
    city={Nanjing, Jiangsu},
    % citysep={}, % Uncomment if no comma needed between city and postcode
    postcode={210093}, 
    % state={},
    country={China}}
\cortext[cor1]{Corresponding author}

% \cortext[]{\textsuperscript{\dag}~Haichuan Hu and Congqing He are co-first authors.}

\fntext[1]{Haichuan Hu and Congqing He are co-first authors.}

\begin{abstract}
Large Language Models (LLMs) have significantly advanced various NLP tasks, yet they still suffer from hallucinations, a phenomenon where LLMs generate plausible yet counterfactual, irrelevant, or logically erroneous responses. 
Recently, Retrieval-Augmented Generation (RAG) enhances the in-context learning abilities of LLMs by incorporating external knowledge, thereby exhibiting the potential to alleviate hallucinations.
However, hallucinations still persist in RAG due to incomplete knowledge extraction and insufficient reasoning over retrieved content. 
In this paper, we propose LRP4RAG, a novel method based on the Layer-wise Relevance
Propagation (LRP) algorithm for detecting hallucinations in RAG. The key insight of LRP4RAG is to dive into the relevance between the RAG context and the model's output, leveraging characteristics such as relevance distribution and consistency to serve as indicators for hallucination detection.
Specially, we adapt LRP to analyze the relevance between the context and the model's output. 
Based on this, we propose two variants of LRP4RAG, LRP4RAG$_\textrm{Classifier}$ and LRP4RAG$_\textrm{LLM}$. LRP4RAG$_\textrm{Classifier}$ converts the relevance matrix into feature vectors for classification. LRP4RAG$_\textrm{LLM}$ prunes the original context using relevance and prompt-based methods, and then performs consistency checks between the pruned contexts and the answer to determine whether hallucination exists. 
We extensively evaluate LRP4RAG against 10 state-of-the-art baselines on two benchmarks and six metrics. 
Experimental results show that LRP4RAG outperforms all baselines, e.g., improving the most recent baseline SEP by 4.1\% and 5.1\% in terms of accuracy.
To the best of our knowledge, LRP4RAG is the first work to apply additive interpretability to hallucination detection in RAG.
Our work further provides new insights into the causes and characteristics of hallucinations in RAG from a relevance-based perspective.
\end{abstract}


\maketitle

\section{Introduction}

Large Language Models (LLMs) have significantly advanced the field of Natural Language Processing (NLP). 
However, they are prone to generating plausible yet factually incorrect answers, a phenomenon commonly referred to as "hallucination". 
The issue of hallucination has been observed across various NLP tasks~\citep{DBLP:journals/corr/abs-2401-11817,DBLP:conf/eacl/GuerreiroVM23, DBLP:conf/acl/WangCVFMR23, DBLP:conf/acl/JiLLYWZF23, DBLP:journals/tacl/AdlakhaBLMR24, he2023explaining}, posing serious challenges to the reliability and safety of LLMs when deployed in practical scenarios.
Thus, researchers have proposed different approaches to detect hallucinations in LLMs, including
(1) uncertainty-based methods~\citep{DBLP:conf/emnlp/ManakulLG23,chen2024insidellmsinternalstates,DBLP:conf/emnlp/FadeevaVTVPFVGP23,DBLP:conf/emnlp/ZhangQGDZZZWF23,DBLP:journals/corr/abs-2403-04696}, which estimate the confidence of model outputs by analyzing internal states;
 (2) consistency-based methods~\citep{DBLP:conf/iclr/0002WSLCNCZ23}, which apply perturbations to the original inputs and then evaluate the self-consistency of the outputs; and (3) task-specific methods~\citep{DBLP:conf/emnlp/QiuZKPC23,guerreiro2023hallucinations}, which focus on specific NLP tasks (e.g., machine translation).

Recently, Retrieval-Augmented Generation (RAG)~\citep{DBLP:conf/nips/LewisPPPKGKLYR020} has demonstrated potential in mitigating  LLM hallucinations by enhancing their in-context learning capabilities through the incorporation of external knowledge.
Nevertheless, hallucinations may still arise within RAG pipelines~\citep{magesh2024hallucinationfreeassessingreliabilityleading,wu2023ragtruth}. In our study, we mainly investigate two specific causes of hallucinations in RAG:

\begin{figure}[t]
  \centering
  \includegraphics[width=0.5\textwidth]{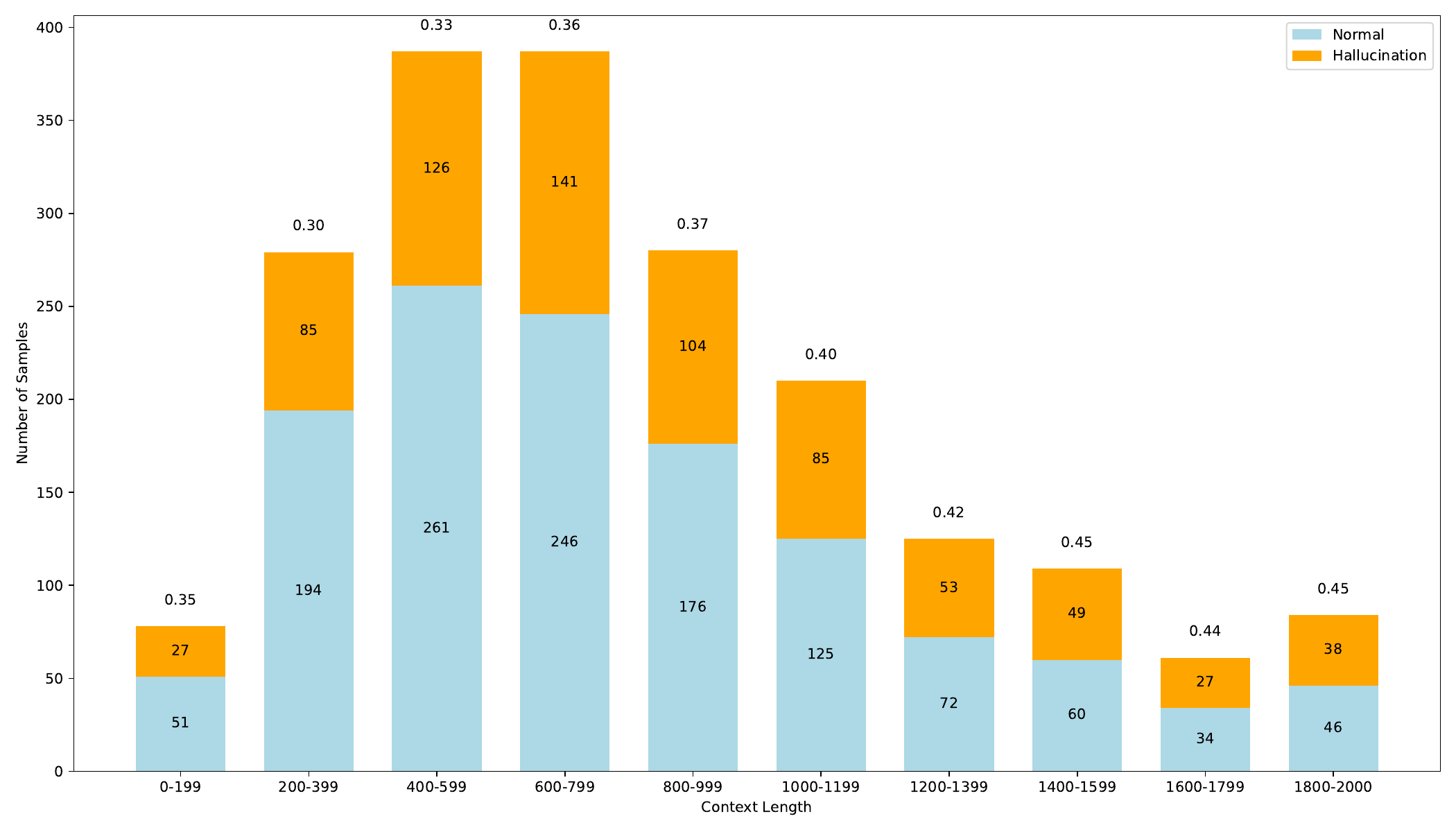}
  \caption{Normal/hallucination distribution sorted by context length collected from Dolly-15k dataset with 2000 samples (generated by Qwen2.5-3B, 1265 normal, 735 hallucinations).}
  \label{fig:distribution}
\end{figure}

\textbf{(1) Long context}.
The first cause of RAG hallucination is the excessively long context.
Recent studies such as “Lost in the Middle”~\citep{DBLP:journals/tacl/LiuLHPBPL24} highlight that LLMs often fail to identify the most relevant information when dealing with lengthy contexts. Context redundancy hinders the generator from outputing correct answers. To validate this, we analyze 2,000 samples with context lengths less than 2,000 tokens from the Dolly-15k dataset~\citep{DatabricksBlog2023DollyV2} and use Qwen2.5-3B as the generator.
As shown in Figure~\ref{fig:distribution}, the hallucination rate generally increases with the context length, reaching its highest point of around 45\% when the context length approaches 2000.
This observation suggests that overly long contexts can lead to hallucinations in RAG, and motivates us to explore context pruning as a way to detect and reduce hallucinations in RAG.

\begin{figure*}[t]
  \centering
  \includegraphics[width=\textwidth]{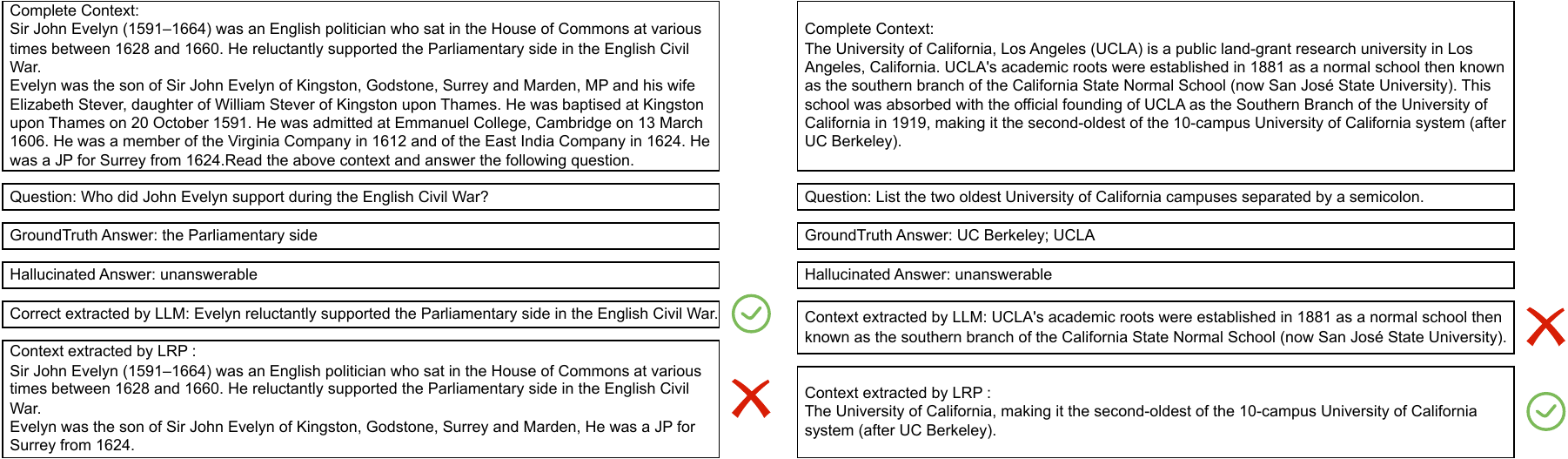}
  \caption{Examples illustrating the mismatch between explicit evidence and implicit evidence.}
  \label{fig:motivation}
\end{figure*}

\textbf{(2) Mismatch between words and thoughts.} 
Another cause of RAG hallucination is the inconsistency between the model's output and its internal thoughts.
When explicitly asked to identify which parts of the context are most relevant to a given question, an LLM is typically able to provide a response.
We refer to this response as the model's "words", which intuitively represents the explicit evidence of the model's response.
However, A natural question thus arises: \textit{does these words truly reflect the model’s internal thoughts?}
To explore this, we apply Layer-wise Relevance Propagation (LRP)~\citep{DBLP:journals/corr/BinderMBMS16} to trace which parts of context contribute most to the model's answer, referred to as the model's implicit "thoughts".
We then compare the model's explicit "words" against implicit "thoughts". Frequent inconsistencies between the two perspectives are found to be associated with the occurrence of hallucinations.
To demonstrate this phenomenon intuitively, we present two motivating examples from the Dolly-15k dataset, as shown in Figure~\ref{fig:motivation}.
As the example on the left shows, the LLM is able to correctly extract the most relevant context which contains the correct answer "Evelyn". 
However, during the RAG process, we find that the LLM's internal thoughts diverge significantly from the correct answer, leading to failure in answering "Who did John Evelyn support during the English Civil War?". 
The example on the right illustrates the opposite case.
Although the LLM is unable to directly extract the correct context, its internal thoughts are correct during the RAG process. The context extracted through LRP includes the two oldest schools in California. 
Despite this, RAG still fails to output the correct answer, indicating that even though the model's internal thoughts are correct, it may still fail to express the answer accurately.

Motivated by these observations, we propose LRP4RAG, a novel hallucination detection approach that leverages LRP to perform context pruning and consistency analysis in RAG.
Specifically, the traditional LRP algorithm~\citep{DBLP:journals/corr/BinderMBMS16} is customized for transformer-based LLMs to analyze relevance between context and answer. 
During context pruning, only the most relevant parts of the context are retained.
We also prompt LLMs to identify the most critical context and then compare the consistency of the two versions of contexts to determine whether hallucination occurs. 
We further check for any contradictions between the pruned contexts and the answer. 
Finally, we employ  LLMs and binary classifiers to detect hallucination based on the the above evidence.

We conduct extensive experiments to compare LRP4RAG with ten state-of-the-art LLM hallucination detection approaches  (including both uncertainty-based and consistency-based ones) on two widely adopted benchmarks (RAGTruth~\citep{wu2023ragtruth}, Dolly-15k~\citep{DatabricksBlog2023DollyV2}) across six evaluation metrics. 
The experimental results demonstrate that LRP4RAG outperforms all existing LLM hallucination detection approaches with an accuracy of 77.2\% and
76.2\%, improving the most recent baseline SEP~\citep{kossen2024semanticentropyprobesrobust} by
4.1\% and 5.1\%.

Our contributions can be summarized as follows: 
\begin{itemize}
    \item We highlight the underexplored challenge of hallucination in RAG pipelines and emphasize the need for effective detection strategies. 
    To the best of our knowledge, LRP4RAG is the first work to apply additive explainability to RAG hallucination detection.
    \item We analyze the correlation between RAG hallucination and context-to-answer relevance, revealing the underlying characteristics and previously unrecognized causes of RAG hallucination.
    \item We propose LRP4RAG, an LRP-based hallucination detection method combining context pruning and consistency check together, which outperforms all state-of-the-art baselines on two public benchmarks and six metrics.
\end{itemize}

\section{Background}
% \subsection{Background}
Our work is grounded in additive interpretability~\citep{agarwal2021neuraladditivemodelsinterpretable,10.1007/978-3-030-28954-6_10,lundberg2017unifiedapproachinterpretingmodel}, which aims to decompose the model's prediction into the sum of contributions from each input feature.  In this paper, we focus on transformer-based LLMs (e.g., GPT, Llama, Qwen) due to their widespread use. Among various additive explanation methods, we choose Layerwise Relevance Propagation (LRP)~\citep{Bach2015OnPE}. Compared to peer methods like Shapley~\citep{lundberg2017unifiedapproachinterpretingmodel}, DeepLIFT~\citep{shrikumar2019learningimportantfeaturespropagating}, and Gradient × Input~\citep{simonyan2014deepinsideconvolutionalnetworks}, LRP provides better local consistency, computational efficiency, and can effectively handle nonlinear relationships.

\subsection{Layer-wise Relevance Propagation}
Layer-wise Relevance Propagation (LRP)\citep{Bach2015OnPE} is an explainability algorithm used to calculate the relevance between the model's inputs and outputs, as well as between the layers.

LRP assumes that a function $f_j$ with $N$ input neurons $x = \{x_i\}_{i=1}^N$ can be decomposed into individual contributions of single input neurons $R_{i\leftarrow j}$ (called ``relevances"). Here, $R_{i\leftarrow j}$ denotes the amount of output $j$ that is attributable to input $i$. LRP is formulated as:

 \begin{equation}
    R_i = \sum_j R_{i \leftarrow j}
    \label{eq:aggretation}
\end{equation}

Although the relevance between individual neurons is variable, the total relevance at the layer level remains constant, which follows the \textit{conservation rule}:

\begin{equation}
\label{lrp:conservation}
     R^{l-1} = \sum_i R^{l-1}_i = \sum_{i,j} R^{(l-1, l)}_{i\leftarrow j} = \sum_j R^l_j = R^l
\end{equation}

\subsection{$\varepsilon$-LRP Applied to Linear and Non-linear Layers}

For a linear layer, the forward propagation is defined as: 

\begin{align}
    & z_j = \sum_i \textbf{W}_{ji} \ x_i + b_j \label{eq:linear} \\
    & a_j = \sigma(z_j)
\end{align}

Where \(W_{ji}\) are the weights between layers, each neuron \(a_j\) in the succeeding layer is obtained by summing the weighted inputs from the neurons \(x_i\) of the previous layer, adding the bias \(b_j\), and then applying the activation function \(\sigma\). We follow the $\varepsilon$-LRP \citep{Bach2015OnPE} rule to compute the relevance for linear layers. $\varepsilon$ is a very small constant used to ensure that the denominator is not zero.
\begin{equation}
    R_i^{l-1} = \sum_j \textbf{W}_{ji} x_i \frac{R_j^l}{z_j(\textbf{x}) + \varepsilon}
    \label{eq:epsilon}
\end{equation}

For a non-linear layer, we adopt an LRP rule similar to the $\varepsilon$-LRP \citep{Bach2015OnPE} rule, as shown in Equation~\ref{eq:assignment}. The difference lies in that: (1) $\textbf{W}_{ji}$ is replaced by $\textbf{J}_{ji}$. The Jacobian $\textbf{J}$
is used to calculate the relevance of neurons between layers. Specifically, $\textbf{J}_{ji}$  represents the association between the \( j \)-th neuron in the \( l \)-th layer and the \( i \)-th neuron in the \( (l-1) \)-th layer. (2) $z_j$ is replaced by $f_j$. We substitute the layer function with its first-order Taylor expansion to facilitate numerical computation.
\begin{equation}
    R_i^{l-1} = \sum_j \textbf{J}_{ji} \ x_i \frac{R_j^{l}}{f_j(\textbf{x}) + \varepsilon }
    \label{eq:assignment}
\end{equation}

\subsection{LRP vs Attention Heads}
\begin{figure}[t] %这里使用的是强制位置，除非真的放不下，不然就是写在哪里图就放在哪里，不会乱动

	\centering  %图片全局居中
	% \vspace{-0.4cm} %设置与上面正文的距离
	\subfigtopskip=2pt %设置子图与上面正文或别的内容的距离
	\subfigbottomskip=2pt %设置第二行子图与第一行子图的距离，即下面的头与上面的脚的距离
	\subfigcapskip=2pt %设置子图与子标题之间的距离
	\subfigure[Girl]{
		\includegraphics[width=0.3\linewidth]{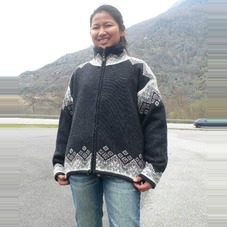}
        \label{sub-1}
        }
	\subfigure[Resnet50]{
		\includegraphics[width=0.3\linewidth]{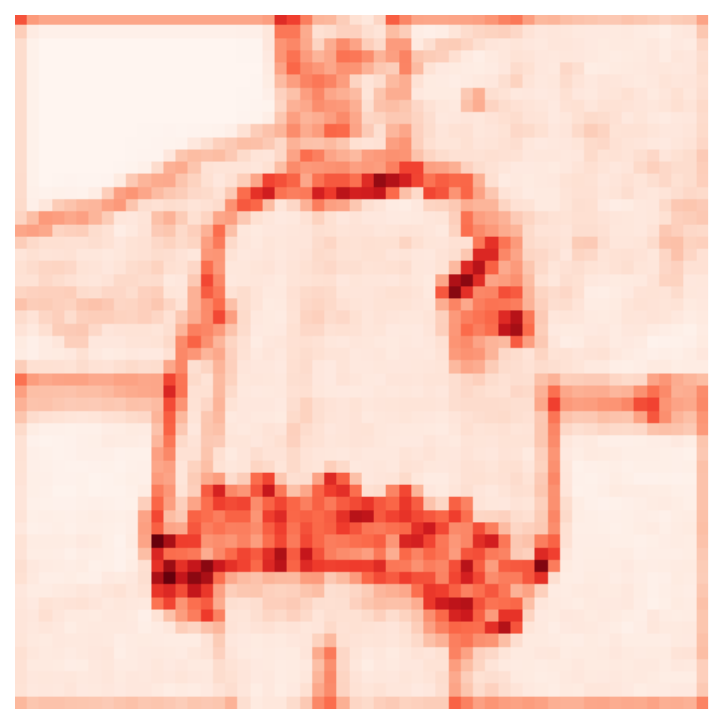}
        \label{sub-2}
        }
        \subfigure[Resnet50-LRP]{
		\includegraphics[width=0.3\linewidth]{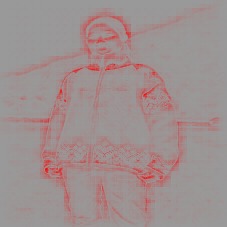}
        \label{sub-3}
        }
        \caption{Example from the LRP paper~\citep{Bach2015OnPE}.}
	
\end{figure}

We find that LRP can make the important features of the model more prominent. As shown in Figure~\ref{girl}, Figure~\ref{sub-1} is an image provided in the original LRP paper~\citep{Bach2015OnPE}, Figure~\ref{sub-2} is the feature map of the image when classified using ResNet50, and Figure~\ref{sub-3} is the feature map after applying the LRP algorithm, also using ResNet50. It can be seen that after applying the LRP algorithm, the important features of the image are retained, while irrelevant features are denoised. Similarly, in the NLP domain, we observe the same pattern: LRP amplifies the key information in the text and filters out the lengthy context.

\label{girl}

Attention Heads-related methods~\citep{sun2025redeepdetectinghallucinationretrievalaugmented,olsson2022incontextlearninginductionheads,ferrando2024informationflowroutesautomatically,he2025simulating} can also serve to filter and highlight important features, which are applicable to transformer-based models and involve analyzing the distribution of attention head weights across different layers to explain the internal mechanisms of the model.

However, due to the lack of clear interpretability of attention heads and the difficulty in determining the importance of attention heads across different layers, we decide to abandon the direct analysis of the model's attention. Instead, we implicitly leverage the attention through the backpropagation of LRP and remove unimportant parts of the attention by filtering the backpropagated features. Compared to directly analyzing attention heads, LRP can be mapped one-to-one with input features, providing greater interpretability and intuition.

\begin{figure*}[t]
  \centering
  \includegraphics[width=\textwidth]{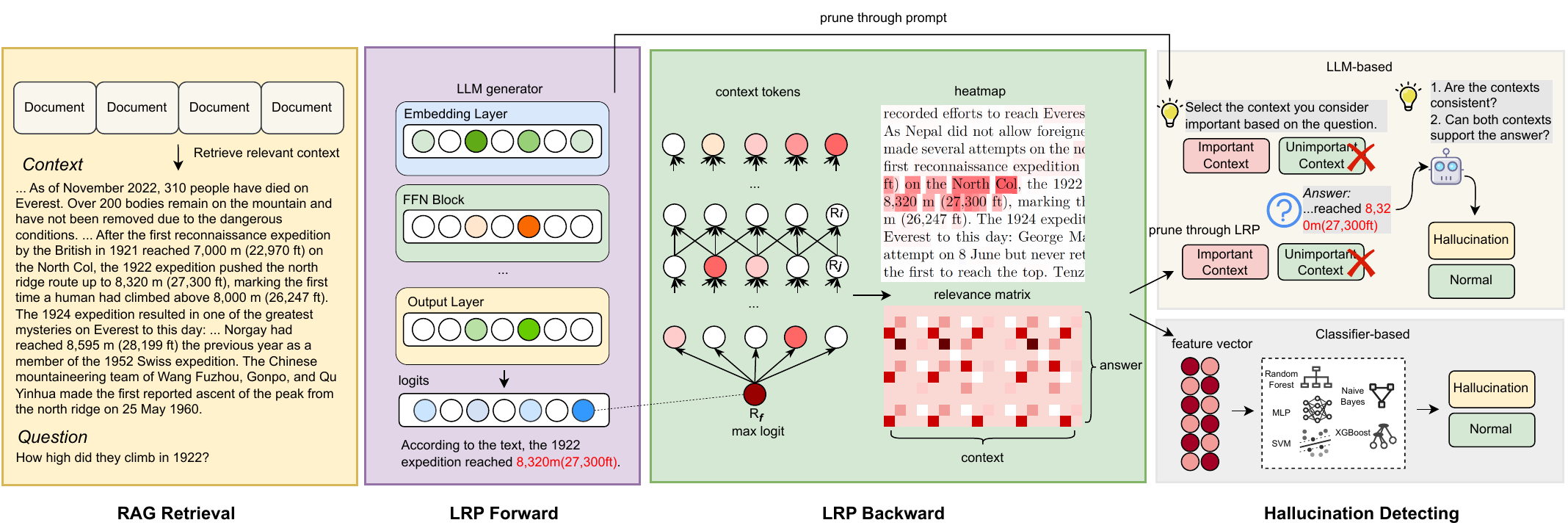}
  \caption{The architecture of our proposed method, LRP4RAG. During RAG Retrieval, the retriever retrieves the most relevant document as context of the question. During LRP Forward, the generator stores the gradients while generating the answer. During LRP Backward, LRP4RAG performs backward propagation through the generator based on predefined LRP rules. During Hallucination Detecting, LRP4RAG$_\textrm{Classifier}$ uses classifiers to perform hallucination detection on the relevance matrix produced by LRP Backward, while LRP4RAG$_\textrm{LLM}$ detects hallucinations through context pruning and consistency checks. } ~\label{fig:overview}
\end{figure*}

\section{Approach}
Figure~\ref{fig:overview} illustrates the overview of our method.
First, given an input query, RAG employs a retriever to extract relevant context from the knowledge base and then formats the prompt based on the query and retrieved context.
Second, during the generation phase, LRP forward is carried out layer by layer through the generator model, with gradients being computed and stored for each individual neuron throughout the procedure. 
Third, after finishing LRP forward, we perform  LRP backward  based on predefined LRP rules, starting from the output layer and propagating backward through the layers to the input layer. LRP backward produces a relevance matrix $\textbf{R}$, which quantifies the contribution of each context token to the generated answer.
In the hallucination detection phase, we utilize two strategies based on $\textbf{R}$. 
The first strategy is classifier-based, where the relevance matrix is transformed into feature vectors and fed into binary classifiers.
The second strategy is LLM-based, where key context segments identified via $\textbf{R}$ are extracted, and the LLM is prompted to self-assess and generate the most relevant context.
We then instruct the LLM to perform consistency checks between these two versions of the context and the answer, and provide a final judgment on whether hallucination has occurred.

\subsection{Task Formulation}
Let $D = {(Q_i, A_i)}_{i=1}^{|D|}$ be a RAG dataset consisting of $|D|$ query-answer pairs $(Q_i, A_i)$, where $Q_i$ is the $i$-th query and $A_i$ is the corresponding answer. Assume that we have a knowledge base containing a large collection of context chunks $C = \{C_j\}_{j=1}^{|C|}$, where $C_j$ denotes the $j$-th context piece (e.g., a passage or document).
Given a query $Q_i \in D$, a retriever, parameterized by $\phi$, first retrieves the most relevant context $C_j$ from the external knowledge base, and then a generator, parameterized by $\theta$, produces an answer $A_i$ based on $C_j$ and $Q_i$. The complete process of RAG is defined as follows:
\begin{equation*}
\underbrace{P(A_i|Q_i)}_{RAG}=\underbrace{P_{\phi}(C_j|Q_i)}_{\text{Retriever}} \underbrace{P_{\theta}(A_i|C_j, Q_i)}_{\text{Generator}}
\end{equation*}
\begin{equation*}
P_{\theta}(A_i|C_j, Q_i) = \prod_{k=1}^{n}P_{\theta}
(a_k|a_1, \cdots, a_{k-1}; C_j; Q_i) 
\end{equation*}
where $\{a_1, \dots, a_{k-1}\}$ denotes the sequence generated prior to the $k$-th token and $n$ is the total number of tokens in the target answer $A_i$. 

Next, we incorporate LRP into the generator. During the generation phase, when generating each token $a_k$, the generator also outputs a relevance vector $\vec{r_k}$ that quantifies the relevance between $a_k$ and the context $C_j$. By aggregating all $\vec{r_k}$, we form a relevance matrix $\mathbf{R_i}$. Thus, the generation phase can be extended as follows:
\begin{equation*}
    P_{\theta}(A_i|C_j, Q_i) = P_{\theta}(A_i, \boldsymbol{R_i}|C_j, Q_i)
\end{equation*}
\begin{equation*}
P_{\theta}(A_i, \boldsymbol{R_i}|C_j, Q_i)
=\prod_{k=1}^{n}P_{\theta}
(a_k, \vec{r_k}|a_1, \vec{r_1}, \cdots, a_{k-1}, \vec{r_{k-1}}; C_j; Q_i) 
\end{equation*}

Finally, a hallucination detector, parameterized by $\gamma$, is introduced to determine whether the generated answer $A_i$ contains hallucinations. After receiving $Q_i$ and its associated context $C_j$, the generator outputs both the answer $A_i$ and the relevance matrix $\mathbf{R_i}$. The hallucination detector then evaluates the answer based on $C_j$, $Q_i$, $A_i$, and $\mathbf{R_i}$, yielding a label $L_i$ that indicates the presence or absence of hallucinations. Formally, this is defined as follows:
\begin{equation*}
    \underbrace{P(L_i|C_j, Q_i)}_{\text{LRP4RAG}} = P_{\theta}(A_i, \boldsymbol{R_i}|C_j, Q_i) \underbrace{P_{\gamma} (L_i|C_j, Q_i, A_i, \boldsymbol{R_i})}_{\text{Hallucination\,Detector}}
\end{equation*}

\subsection{LRP4RAG Framework}
We propose LRP4RAG, a framework for hallucination detection in RAG that leverages LRP to analyze the attribution between retrieved context and generated answers. Given  query $Q_i$ and the retrieved context $C_j$, the LLM generates an answer $A_i$. Hallucinations arise when $A_i$ is inconsistent with $C_j$.
LRP4RAG first applies LRP to derive a relevance distribution $R$, identifying which parts of $C_j$ most influence the generation of $A_i$. Based on $R$, the framework is implemented with two detection strategies:
(1) a classifier-based approach that converts the relevance matrix into feature vectors for binary classification, and
(2) an LLM-based consistency check which compares the model’s internal evidence (via LRP) with its explicit evidence (via prompt) and the output.
The following sections detail the LRP computation and both detection methods.

\subsubsection{LRP for Transformer-based LLMs}
\noindent\textbf{LRP Rules for Transformer.}
This section begins with an overview of various LRP propagation rules tailored for Transformer-based LLMs, which are employed to calculate the relevance across different layers and individual neurons.
To obtain token-level relevance in an LLM, we propagate output relevance backward through each layer of the transformer.  To adapt standard LRP rules for transformer architectures, we adopt an $\epsilon$-LRP formulation, where for each layer we use the local Jacobian $J_{ji} = \partial z_j / \partial x_i$ (the gradient of the layer’s output $z_j$ with respect to its input $x_i$) as the weight for distributing relevance.
In a multi-head attention block, let $\textbf{Q}$, $\textbf{K}$, and $\textbf{V}$ denote the query, key, and value matrices, respectively. The scaled dot-product attention computes attention weights $\textbf{A}$ and output values $\textbf{O}$ as:
% For LLMs with the transformer architecture, the most core component is the non-linear attention, whose forward propagation can be represented as:
\begin{align}
   & \textbf{A} = \text{softmax}\left(\frac{\textbf{Q} \cdot \textbf{K}^\top}{\sqrt{d_k}}\right)~, \\
   & \textbf{O} = \textbf{A} \cdot \textbf{V}~.
   \label{eq:attention}
\end{align}

Where (·) denotes matrix multiplication, \textbf{K} is the key matrix, \textbf{Q} is the queries matrix, \textbf{V} is the values matrix, and \textbf{O} is
the final output of non-linear attention. $d_k$ indicates the embedding dimensions. The computation of attention includes two basic operators: softmax and matrix multiplication. Softmax is represented as:
\begin{equation}
    \text{softmax}_j(\textbf{x}) = \frac{e^{x_j}}{\sum_i e^{x_i}}~.
\end{equation}

The derivative of the softmax has two cases, which depend on the output and input indices $i$ and $j$:
\begin{equation}
    \frac{\partial s_j}{\partial x_i} = 
    \begin{cases}
      s_j(1-s_j) & \text{for } i=j \\
        -s_j s_i & \text{for } i \neq j
    \end{cases}~. \label{eq-9}
\end{equation}

For the two cases listed in Equation~\ref{eq-9}, we bring their partial derivative expressions back into Equation~\ref{eq:assignment} respectively. As shown in Equation~\ref{eq-10}, case(i) represents relevance from output $j$ attributed to input $i \neq j$, while case (ii) represents relevance from output $j$ attributed to input $i = j$.
\begin{equation}
    R_{i\leftarrow j}^{(l-1, l)} = 
    \begin{cases}
       (x_i - s_i x_i) \ R^{l}_i & \text{for } i=j \\
       -s_i x_i \ R^{l}_j & \text{for } i \neq j
    \end{cases}~. \label{eq-10}
\end{equation}

We further apply Equation~\ref{eq-10} to Equation~\ref{lrp:conservation} to obtain the LRP rule for the softmax function: 
\begin{equation}
    R^{l-1}_i = x_i (R^{l}_i - s_i \sum_j R^{l}_j)~,
    \label{eq:lrp-softmax}
\end{equation}
where $R^l_j$ is the relevance of output neuron $j$, and $R^{(l-1)}_i$ is the resulting relevance of input $i$.
For matrix multiplication $\textbf{C} =\textbf{A} \cdot \textbf{B}$ , relevance is calculated for both matrix multiplication terms $\textbf{A}$ and $\textbf{B}$. 
\begin{equation}
R_\textbf{A}=R_\textbf{C} \cdot \textbf{B}^\top \odot \textbf{A}~, \qquad R_\textbf{B}=R_\textbf{C} \cdot \textbf{A}^\top \odot \textbf{B}~,
\end{equation}
where $\odot$ denotes the Hadamard product.

\noindent\textbf{LRP Forward/Backward.} During the generation phase, we perform a forward pass through the LLM while recording the gradients required by the LRP rules described above. Once the model generates a token $a_i$, we initiate LRP backward propagation from the output layer to the input layer.
\begin{equation}
    \overset{Output Layer}{R^{\text{logit}} \xrightarrow{}} \cdots  \overset{Transformer Block}{R^l \rightarrow R^{l-1}} \cdots \overset{Input Layer}{\xrightarrow{} R^{\text{input}}}~.
\end{equation}

The relevance score of the output layer, denoted as $R_{\text{logit}}$, is determined by the maximum logit among the output token’s prediction scores.
\begin{equation}
R^{\text{logit}} = max\,(\text{logits})~.
\end{equation}

Starting with this output-layer relevance, we propagate backwards through each transformer block using the LRP rules, until reaching the input layer. This yields a relevance score for every input token. The result is a relevance matrix  that quantifies how much each context token contributes to the generated answer. We denote by $R_{\text{token},i}$ the relevance score for the $i$-th context token.

\subsubsection{Classifier-based  Hallucination Detection}
In the previous section, we introduce the rules and process of LRP in LLMs. Based on the relevance matrix generated by LRP, we propose a classifier-based hallucination detection method LRP4RAG$_\textrm{Classifier}$.
LRP4RAG$_\textrm{Classifier}$ processes the relevance matrix into feature vectors, which are then used in binary classification to detect hallucinations.
Specifically, the context-to-answer relevance matrix $R$ is convert  into a set of numerical features that capture the interaction between the context and the answer.
We focus on the aggregated per-token relevance scores $R_{\text{input}}$ for the context. Since tokens are subword units after tokenization, we first aggregate token-level relevance to the word level.
If a context word is tokenized into pieces ${t_i}$, we define its word-level relevance as the sum of its token relevances:
\begin{equation}\label{eq-17}
    R_{\text{word}} \;=\; \sum_{i \in \text{tokenize}(\text{word})} R_{\text{token},i} ~,
\end{equation}
which combines all subword contributions for that word. The sequence of relevance scores is then transformed into a feature vector (e.g., by pooling or bucketing the $R_{\text{word}}$ sequence to a fixed length).
This vector is fed into a binary classifier trained to detect hallucinations.
Experiments conducted with a range of classifiers, including Random Forest, Support Vector Machine, MLP, and LSTM.

\subsubsection{LLM-based Hallucination Detection}
In addition to LRP4RAG$_\textrm{Classifier}$, we further propose LRP4RAG$_{\text{LLM}}$, a training-free hallucination detection method that leverages the self-checking mechanism of LLMs during the RAG process. The method is designed to verify whether the \emph{internal evidence} (i.e.,  the context implicitly marked important due to high relevance scores) aligns with the \emph{explicit evidence} (i.e., the context the model explicitly deems most relevant) and the \emph{final answer}. Specifically, the method comprises three stages: Internal Evidence Extraction, Explicit Evidence Extraction, and Consistency Verification.

\noindent\textbf{Internal Evidence Extraction.} 
We use LRP to identify the most relevant parts of the original context when generating the answer.
By backpropagating relevance scores from the output layer to the input layer, we obtain a per-token relevance distribution, which can then be aggregated at the word or sentence level.
Word-level relevance $R_{\text{word}}$ is computed by summing token-level relevance scores for each word, as defined in Equation~\ref{eq-17}.
Next, we identify sentence boundaries in the context and calculate sentence-level relevance by averaging the relevance scores of the words in each sentence.
Formally, if a sentence contains words $\{w_1, \dots, w_N\}$, it is define as:
\begin{equation}
    R_{\text{sentence}} = \frac{1}{N}\sum_{w \in \text{sentence}} R_{\text{word},\,w} ~,
\end{equation}
which assigns each sentence the mean relevance of its constituent words. We then rank all sentences by $R_{\text{sentence}}$ and select the top-$k\%$ as the key context $C_{\text{key}}$, representing what the model implicitly found most relevant.
(In our experiments, $k = 20$ is used.) The set $C_{\text{key}}$ serves as the model’s internal evidence for its answer.
\begin{equation}\label{eq-19}
    \textbf{\textit{C}}_{\text{key}} = \{ s \mid R_{sentence, s} \in \text{top}_{k\%}(R_{sentence}, k) \} ~.
\end{equation}

\noindent\textbf{Explicit Evidence Extraction.} 
We prompt the LLM to output what it considers to be the most important context based on the answer. The LLM is first instructed to score each sentence according to its importance and relevance to the question, then sort the sentences by their scores, and select the most important ones to construct a concise context.
In this way, the LLM prunes the original context, retaining only the parts it deems directly relevant to the answer. We denote the pruned context as $\textbf{\textit{C}}_{llm}$.
Additionally, we instruct the LLM not to perform any summarization or rewriting to ensure that the context remains faithful to the original and is not distorted during pruning.

\noindent\textbf{Consistency Verification.}

After extracting the internal evidence $C_{\text{key}}$ and the explicit evidence $C_{\text{llm}}$, we compare them with the answer $A_i$ for consistency check. Particularly, we perform three types of checks, listed as follows.

\textbf{First check.} The first check is between \(\textbf{\textit{C}}_{\text{key}}\) and \(\textbf{\textit{C}}_{\text{llm}}\), where we assess whether the key focuses of both are consistent. If the similarity is high and there is significant overlap,  it suggests that the model’s internal reasoning aligns with the basis for its response.
Conversely, if \(\textbf{\textit{C}}_{\text{key}}\) and \(\textbf{\textit{C}}_{\text{llm}}\) differ significantly,  it indicates a discrepancy between the model’s internal reasoning and its expressed justification, which is likely to lead to hallucinations. Additionally, we use Sentence-Transformer to compute semantic similarity to help the LLM determine the consistency between \(\textbf{\textit{C}}_{\text{key}}\) and \(\textbf{\textit{C}}_{\text{llm}}\), as shown in Equation~\ref{eq-20}. 
\begin{equation} \label{eq-20}
    Consistency_1 = sim \,(Embed\,(\,\textbf{\textit{C}}_{\text{key}}\,), Embed\,(\,\textbf{\textit{C}}_{\text{llm}}\,)) ~.
\end{equation}

\textbf{Second check.} The second check is between \(\textbf{\textit{C}}_{\text{key}}\) and the answer $A_i$, where we prompt the LLM to evaluate whether the answer contradicts \(\textbf{\textit{C}}_{\text{key}}\), as shown in Equation~\ref{eq-21}. The correct answer should be logically consistent with both the original context and the pruned context. If any contradictions arise, it indicates that the model produces a hallucination. 
\begin{equation} \label{eq-21}
    Consistency_2 = Consistency\_Check \,(\textbf{\textit{C}}_{\text{key}}, A_i) ~.
\end{equation}

\textbf{Third check.} The third check is between \(\textbf{\textit{C}}_{\text{llm}}\) and the answer $A_i$, where we prompt the LLM to assess for semantic consistency in the same way as in the second check, as shown in Equation~\ref{eq-22}. After completing these three checks, we require the LLM to self-assess whether its response contains any hallucinations based on $Consistency_1, Consistency_2, Consistency_3$.
\begin{equation} \label{eq-22}
    Consistency_3 = Consistency\_Check \,(\textbf{\textit{C}}_{\text{llm}}, A_i)  ~.
\end{equation}

\section{Experimental Setup}
\subsection{Datasets}
We conduct experiments on RAGTruth ~\citep{wu2023ragtruth} and Dolly-15k~\citep{DatabricksBlog2023DollyV2}. RAGTruth consists of 989 human-annotated RAG samples collected from different LLMs. In particular, Llama-7B includes 510 hallucinated samples and 479 normal samples, while Llama-13B includes 399 hallucinated samples and 590 normal samples. Dolly-15k consists of 15000 samples across six scenarios (closed\_qa, open\_qa, information\_extraction, classification, brainstorming, summarization). We filter it to meet the RAG conditions with the following rules: the context is non-empty and has a length of less than 2000 characters. We then use GPT-4 to compare the model's output with the standard answers to label the dataset. Finally, to ensure a balanced ratio of positive and negative samples, we select 2000 samples as our test set.

\subsection{Implementation Details} 
 In our study, both LRP4RAG$_\textrm{Classifier}$ and LRP4RAG$_\textrm{LLM}$ are implemented with Pytorch. For LRP4RAG$_\textrm{Classifier}$, we choose 4  classifiers, including Random Forest~\citep{10.1023/A:1010933404324}, SVM~\citep{10.1109/5254.708428}, LSTM~\citep{10.1162/neco.1997.9.8.1735}, and MLP. We use the Adam optimizer to train MLP and LSTM, with a learning rate of 2e-5, a batch size of 64, and a training duration of 50 epochs. The hidden layer size and number of layers of LSTM are set to 256 and 2, respectively. For Random Forest and SVM, we retain the default settings of Pytorch. For LRP4RAG$_\textrm{LLM}$, we load the pre-trained parameters from Hugging Face using the Transformers library and use vLLM 0.6.2 for model inference. The temperature is set to 0, the maximum input length is set to 8000, and the maximum output length is set to 2048.
 
 % On RAGTruth, we use Llama-7b and Llama-13b as RAG generators.On Dolly-15k, we use Qwen-3b and Qwen-7b ~\citep{qwen2025qwen25technicalreport} as RAG generators. We use the same Qwen models for LLM-based LRP4RAG.

\subsection{Evaluation Metrics}

We evaluate the performance of LRP4ARG by comparing the predicted labels with the ground truth labels, which can be categorized into four scenarios: True Positive (TP), False Positive (FP), False Negative (FN), and True Negative (TN).
In particular, TP refers to a normal sample that is correctly identified as normal; FP refers to a hallucinated sample that is incorrectly identified as normal; FN refers to a normal sample that is incorrectly identified as hallucinated; and TN refers to a hallucinated sample that is correctly identified as hallucinated.
Then, we select four standard metrics to evaluate LRP4ARG, defined as follows.

$\bullet$ \textit{Accuracy} measures the proportion of correctly reported (whether the sample is normal or hallucinated) samples.
\begin{equation}
\text{Accuracy} = \frac{TP + TN}{TP + TN + FP + FN}
\end{equation}

$\bullet$ \textit{Recall} measures the ratio of correctly identified normal samples over all the real normal samples.
\begin{equation}
\text{Recall} = \frac{TP }{TP + FN}
\end{equation}

$\bullet$ \textit{Precision} measures the proportion of real normal samples over the reported normal samples.
\begin{equation}
\text{Precision} = \frac{TP }{TP + FP}
\end{equation}

$\bullet$ \textit{F1} measures twice the multiplication of precision and recall divided by the sum of them.
\begin{equation}
\text{F1} = 2 * \frac{Precision * Recall}{Precision + Recall}
\end{equation}

$\bullet$ \textit{Area Under Curve (AUC)} measures the area under the receiver operating characteristic curve.
% indicating APCA approaches' ability to distinguish between classes.
Here, $rank\_i$ represents the predicted rank of the $i$-th positive sample, $P$ represents the number of positive samples, and $N$ represents the number of negative samples.
\begin{equation}
\text{AUC} = \frac{\sum_{i=1}^{P}rank_i - \frac{P(P+1)}{2}}{P\times N}
\end{equation}

$\bullet$ \textit{PCC} (Pearson Correlation Coefficient) is a measure of the linear correlation between two variables.

\begin{equation}
\rho_{X,Y} = \frac{\text{Cov}(X, Y)}{\sigma_X \sigma_Y}
\end{equation}
where \(\text{Cov}(X, Y)\) is the covariance of \(X\) and \(Y\), \(\sigma_X\) is the standard deviation of \(X\), \(\sigma_Y\) is the standard deviation of \(Y\).

% \subsection{Experimental Settings}
% During the model generation phase, we keep the model's hyperparameters consistent with the dataset settings. In the LRP phase, to ensure numerical stability, we set epsilon to 1e-6 when normalizing the relevance scores. We use the radial basis function (RBF) kernel in SVM classifier, and set up a two-layered LSTM classifier, with hidden\_size set to 256. The learning rate for the LSTM classifier is set to 5e-4.

\subsection{Baselines}
We compare LRP4RAG with the following baselines.

\textbf{Prompt}~\citep{wu2023ragtruth} We manually craft hallucination detection prompt to instruct LLMs (Llama-7b and GPT-3.5-turbo) to identify RAG hallucinations and corresponding span. 

\textbf{SelfCheckGPT}~\citep{manakul2023selfcheckgpt} We utilize SelfCheckGPT to check consistency between sampled responses. If inconsistency exceeds the threshold, we suppose hallucinations happen.

\textbf{Fine-tune}~\citep{wu2023ragtruth} We fine-tune Llama-7b and Qwen-7b to identify RAG hallucinations.

\textbf{Perplexity}~\citep{ren2023outofdistributiondetectionselectivegeneration} Perplexity measures the uncertainty of an LLM by evaluating how surprised it is with the given text.

\textbf{LN-Entropy}~\citep{malinin2021uncertaintyestimationautoregressivestructured} LN-Entropy adjusts the entropy of a sequence by its length to provide a more consistent measure of uncertainty.

\textbf{Lexical Similarity}~\citep{lin-etal-2022-towards} Lexical Similarity assesses the consistency of generated text by comparing the lexical overlap between different generations of the same input.

\textbf{Energy}~\citep{liu2021energybasedoutofdistributiondetection} Energy is a method for out-of-distribution (OOD) detection that uses the energy function of a neural network to determine the likelihood of an input being in-distribution.

\textbf{EigenScore}~\citep{chen2024insidellmsinternalstates} EigenScore is proposed to better evaluate
responses’ self-consistency, which exploits the eigenvalues of responses’ covariance matrix to measure the semantic consistency/diversity in the dense embedding space. 

\textbf{SAPLMA}~\citep{azaria-mitchell-2023-internal} SAPLMA trains a classifier on LLM activation values to detect hallucinations. It captures internal signals from the LLM’s hidden layers to identify when the model might generate a hallucination.

\textbf{SEP} (Semantic Entropy Probe)~\citep{kossen2024semanticentropyprobesrobust} SEP uses linear probes trained on the hidden states of LLMs to detect hallucinations by analyzing the semantic entropy of the tokens before generation.

% • RQ1: What is the difference between normal and hallucinated samples in RAG?
% • RQ2: How effective is Classifier-based LRP4RAG?
% • RQ3: How effective is LLM-based LRP4RAG?
% • RQ4: How efficient is LRP4RAG?
\section{Results}

We evaluate LRP4RAG on the following research questions:
\begin{itemize}
\item \textbf{RQ1: } What is the difference between normal and hallucinated samples in RAG?
\item \textbf{RQ2: }  How does LRP4RAG compare with state-of-the-art methods in detecting hallucinations?
% 讨论你的方法和别人的方法
% 应该RQ2和RQ3可以共用一个表达
% 你用两个数据集做实验，所以每个表应该都包含  classifer-based  and  llm-based 效果

\item \textbf{RQ3: }  How do LRP4RAG$_\textrm{Classifier}$ and LRP4RAG$_\textrm{LLM}$ differ in performance?
%  可以分析效果以及时间上的性能
\item \textbf{RQ4: }  What is the impact of each component in LRP4RAG?
% 分析消融实验，如果有的话。
\item \textbf{RQ5: }  How do hyper-parameter settings affect LRP4RAG’s performance?
% 把你附录的参数实验拿上来 通过折线图的方式展示就行

% 有case  study 可以再加个例子分析
\end{itemize}

\subsection{RQ1: Hallucination vs Normal —— an Empirical Study on Context-to-answer Relevance Distribution} \label{result-1}
\textbf{Experimental Design. }
In response to RQ1, we conduct an empirical study on 989 RAG samples from RAGTruth. We aim to analyze the differences in the distribution of relevance between hallucinated samples and normal samples, thus demonstrating the reasonableness of employing LRP for detecting RAG hallucinations. 

\begin{figure}[htbp]
\centering
\includegraphics[width=0.5\textwidth]{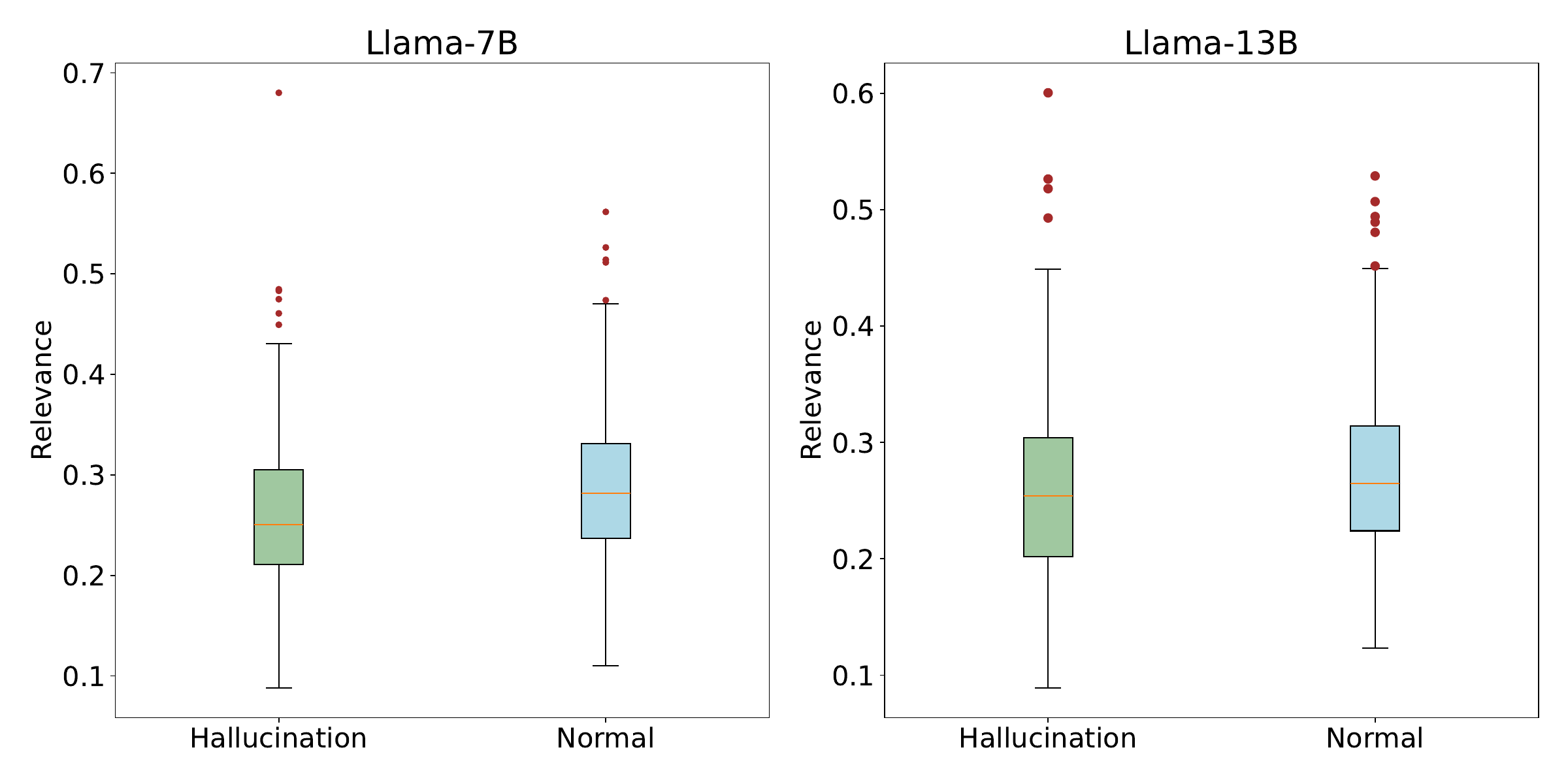} % Replace with your image path
        \caption{Comparison of sample-level relevance between normal and hallucinated samples.}
        \label{fig:box}
\end{figure}

\begin{figure}[htbp]
\centering
\includegraphics[width=0.5\textwidth]{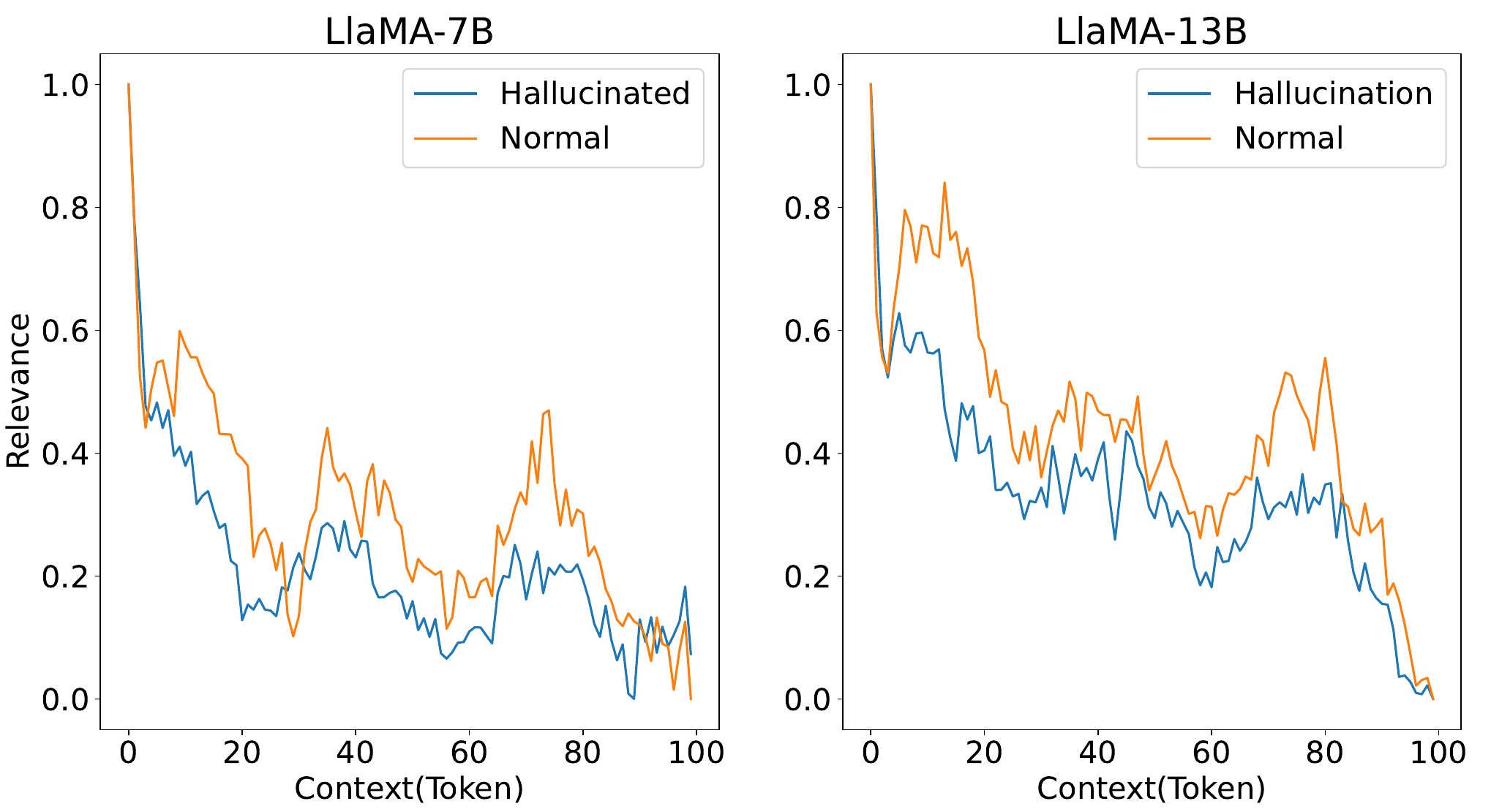} % Replace with your image path
        \caption{Comparison of token-level relevance between normal and hallucinated samples.}
        \label{fig:line}
\end{figure}

\begin{figure}[htbp]
\centering
\includegraphics[width=0.5\textwidth]{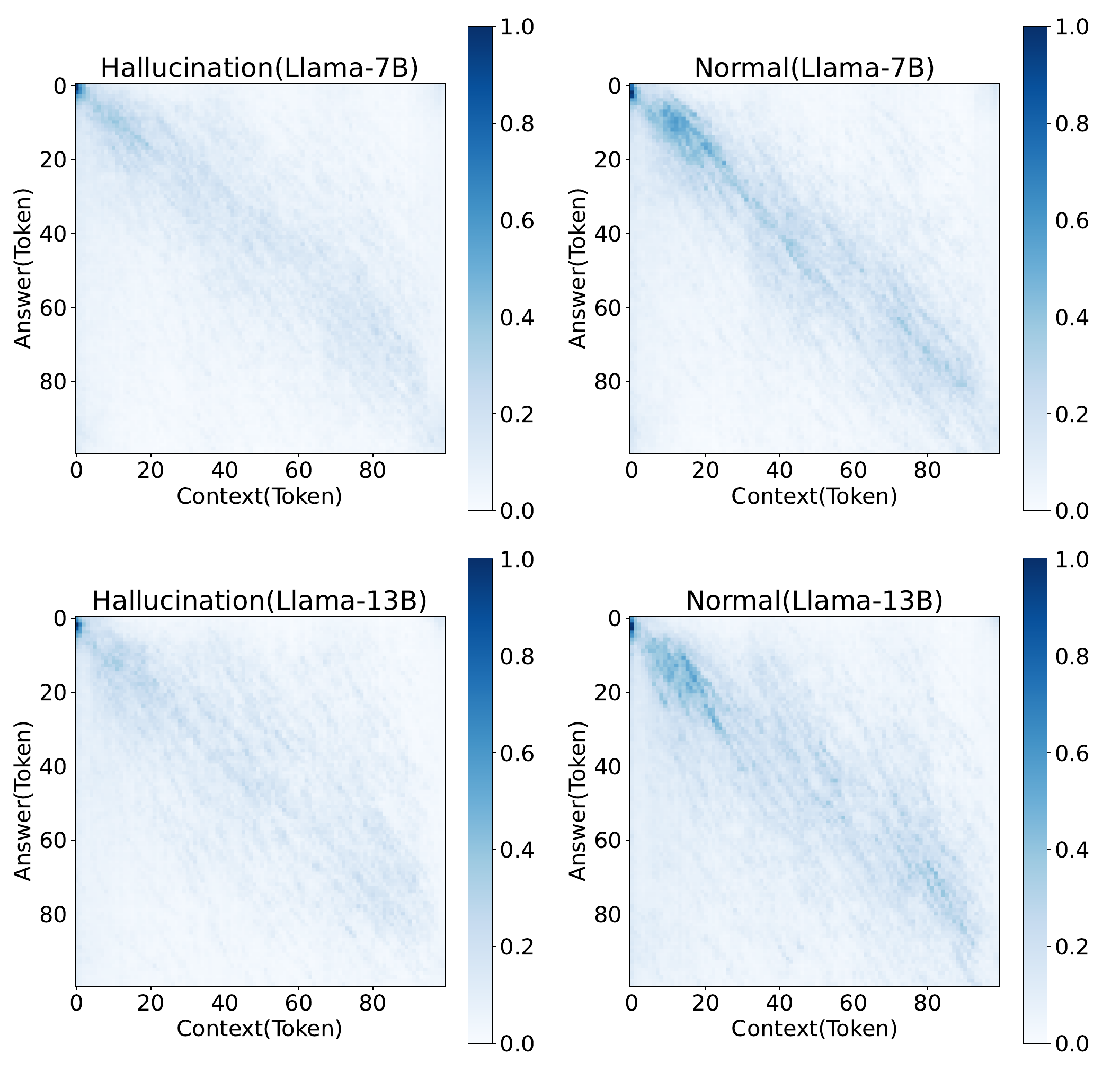} % Replace with your image path
        \caption{Hotmaps of context-to-answer relevance.}
        \label{fig:hotmap}
\end{figure}

\textbf{Results. } As shown in Figure~\ref{fig:box},~\ref{fig:line},~\ref{fig:hotmap}, we illustrate the differences from 3 perspectives (box graph, line graph and heatmap). 

\textbf{Sample-level relevance.} We use box plots in Figure~\ref{fig:box} to visualize the cumulative relevance at sample-level. Excluding few outliers, it can be observed that the median relevance for normal samples  is higher than that for hallucinated samples.   This indicates that, compared to hallucinated samples, normal samples exhibit a stronger correlation between the prompt and the response.

\textbf{Token-level relevance.} We further calculate token-level relevance distribution for normal and hallucinated samples, respectively. The results are plotted in Figure~\ref{fig:line}, where the relevance curve for normal samples consistently lies above that of hallucinated samples.  This indicates that answers in normal RAG samples are more strongly associated with specific tokens in the context, whereas  hallucinated samples tend to weaken the context. These findings suggest that one key reason for hallucination in RAG is the model’s reduced reliance on the provided context, instead relying  more heavily on parameterized knowledge.

\textbf{Context-to-answer relevance.} The heatmaps in Figure \ref{fig:hotmap} show the context-to-answer relevance distribution, which are consistent with Figure~\ref{fig:box} and Figure~\ref{fig:line}. The heatmaps on the right contain more dark regions than the left ones, indicating that stronger relevance between the context and answer exists in normal samples, compared to hallucinated samples . Besides,  the pair of heatmaps in the upper row exhibit more pronounced differences compared to those in the lower row, suggesting that hallucinations in Llama-7B are more  distinguishable than those in Llama-13B. This is because, as the model's capability increases, it learns to better disguise wrong answers, making hallucinations more difficult to detect.

Based on the above observations, we design a threshold-based hallucination detection method to preliminarily validate the effectiveness of LRP in detecting RAG hallucinations. We first normalize the relevance distribution generated by LRP to the range (0,1), and then calculate the mean relevance at the token level. If the mean is greater than or equal to the threshold, the sample is classified as normal;  otherwise, it is classified as hallucinated. Table~\ref{tab:threshold} presents the result under different threshold values. Full results can be found in Appendix \ref{full-threshold}.
For Llama-7B, the threshold value is found to be optimally effective at approximately $t = 0.5$, achieving an accuracy exceeding 60\% and a recall over 80\%. For Llama-13B, the threshold value is around $t = 0.4$, with accuracy slightly below 60\%. Overall, the hallucination detection performance for Llama-13B is lower than that for Llama-7B, indicating that hallucinations in larger parameter models are more difficult to detect.

\begin{table*}[htbp]
\centering

\resizebox{0.9\textwidth}{!}{
\begin{tabular}{c|l|cccc|cccc} 
\toprule
& & \multicolumn{4}{c|}{\textbf{RAGTruth$_\textrm{Llama-7B}$}} & \multicolumn{4}{c}{\textbf{RAGTruth$_\textrm{Llama-13B}$}} \\

\cmidrule{2-10}
\multirow{6}{*}{\rotatebox{90}{$\overline{R^{token}}$}}& \textbf{Relevance Threshold} &   \textbf{Accuracy} & \textbf{Precision} & \textbf{Recall} & \textbf{F1} & \textbf{Accuracy} & \textbf{Precision} & \textbf{Recall} & \textbf{F1}  \\
\cmidrule{2-10}
 & $t = 0.41$  & 55.61\%    &  68.33\%       & 26.03\% & 37.65\% & 59.36\%    & 49.20\%    & 23.54\% & 31.74\%\\
& $t = 0.44$  & 58.34\%    &  63.73\%       & 44.25\% & 52.22\% & 58.44\%    & 48.29\%    & 40.48\% & 43.86\%\\
& $t = 0.47$  & 60.16\%    &  61.00\%       & 61.85\% & 61.38\% & 55.81\%    & 46.30\%    & 60.26\% & 52.22\%\\
& $t = 0.5$   & 60.16\%    &  58.06\%       & 81.42\% & 67.72\% & 51.87\%    & 44.56\%    & 79.09\% & 56.87\%\\
& $t = 0.53$  & 58.24\%    &  55.74\%       & 92.59\% & 69.53\% & 47.62\%    & 43.09\%    & 92.75\% & 58.71\%\\
\bottomrule
\end{tabular}}
\caption{Results of threshold-based RAG hallucination detection method which utilizes LRP to calculate token-level relevance $\overline{R^{token}}$. Set threshold $t$ for $\overline{R^{token}}$. When $\overline{R^{token}}$ < $t$, a sample is identified as hallucination. When $\overline{R^{token}}$ >= $t$, a sample is identified as normal. $\overline{R^{token}}$ is the mean of normalized $R^{token}$.}\label{tab:threshold}
\end{table*}

\subsection{RQ2: Comparison with State-of-the-art Baselines}
\textbf{Experimental Design. }
In RQ2, we aim to evaluate the effectiveness of LRP4RAG in hallucination detection. We compare LRP4RAG with 10 state-of-the-art baselines on RAGTruth and Dolly-15k using 6 metrics.

\textbf{Results.} The comparison results on RAGTruth are listed in Table~\ref{tab:main-results}. Compared to existing methods, LRP4RAG has strong performance across all metrics. On RAGTruth$_\textrm{Llama-7B}$, LRP4RAG$_\textrm{LLM}$ achieves the highest accuracy (73.41\%), precision (71.18\%) and F1-score (73.54\%), which makes an improvement of 11.84\%, 8.68\% and 5.92\% over non-LRP based methods. LRP4RAG$_\textrm{Classifier}$ also shows considerable performance, among which LRP4RAG$_\textrm{SVM}$ achieves an accuracy of 69.16\%, a precision of 69.35\%, and an F1-score of 70.64\%, second only to LRP4RAG$_\textrm{LLM}$. On RAGTruth$_\textrm{Llama-13B}$, LRP4RAG$_\textrm{LLM}$  also outperforms all  baselines, achieving an accuracy of 71.70\%, a precision of 77.14\%, and an F1-score of 75.86\%. Notably, LRP4RAG$_\textrm{Classifier}$ maintains an accuracy around 70\%, but the recall drops to around 50\%, indicating that LRP4RAG$_\textrm{Classifier}$ tends to identify samples as hallucinated, leading to more false negatives.

\begin{table*}[htbp]
\centering

\resizebox{\textwidth}{!}{
\begin{tabular}{l|cccc|cccc} 
\toprule
\multirow{2}{*}{\textbf{Methods}}& \multicolumn{4}{c|}{\textbf{RAGTruth$_\textrm{Llama-7B}$}} & \multicolumn{4}{c}{\textbf{RAGTruth$_\textrm{Llama-13B}$}} \\
\cmidrule{2-9}
 &   \textbf{Accuracy} & \textbf{Precision} & \textbf{Recall} & \textbf{F1} & \textbf{Accuracy} & \textbf{Precision} & \textbf{Recall} & \textbf{F1}  \\
\midrule
Prompt$_\textrm{llama-7b}$                       & 52.38\%                        & 52.64\%                         & 76.08\%& 62.23\%& 49.65\%                        & 41.02\%                         & 56.64\% & 47.58\%\\
Prompt$_\textrm{gpt-3.5-turbo}$ & 54.39\% & 56.91\% & 47.65\%& 51.87\% & 57.84\% & 47.58\% & 44.36\% &45.91\%\\
SelfCheckGPT$_\textrm{llama-7b}$ & 53.79\% & 53.32\% & 83.53\%& 65.09\% & 50.76\% & 43.66\% & 75.94\% &55.44\%\\
SelfCheckGPT$_\textrm{gpt-3.5-turbo}$ & 54.29\%& 53.27\% & \textbf{92.54\%} & 67.62\% &47.93\%& 43.01\% & \textbf{89.47\%} & 58.10\% \\
Fintune$_\textrm{llama-7b}$ & 61.57\%& 62.50\% & 65.75\%& 63.58\% & 64.90\%& 63.70\% & 27.92\% & 37.62\% \\
Fintune$_\textrm{qwen2-7b}$ & 60.46\%& 61.76\% & 64.34\%& 61.90\% & 63.80\%& 63.55\% & 25.93\% & 35.89\% \\
Perplexity & - & - & 51.90\%  & 67.49\% & - & - & 51.90\% & 67.49\% \\
Energy & - & - & 50.57\%  & 66.57\% & - & - & 50.57\% & 66.57\%\\
LN-Entropy & - & - & 53.83\% & 66.55\% & - & - & 53.83\% & 66.55\% \\
LexicalSim & - & - & -& -& -& -& -& - \\
EigenScore & - & - & 74.69\% & 66.82\% & - & - & 67.15\% & 66.37\% \\
SAPLMA & - & - & 50.91\% & 67.26\% & - & - & 50.53\% & 65.29\% \\
SEP & - & - & 74.77\% & 66.27\% & - & - & 65.80\% & 71.59\% \\
\midrule
LRP4RAG$_\textrm{RF}$ (Ours)                  & 60.36\%    &  60.04\%                        &69.19\%&64.21\% & 64.41\%    &  61.92\%                        & 30.35\% & 40.52\%\\
LRP4RAG$_\textrm{MLP}$ (Ours)                 & 63.59\%    &  64.24\%                        &66.01\%& 65.10\% & 65.42\%    &  58.90\%                        & 46.40\% & 51.77\%\\
LRP4RAG$_\textrm{SVM}$ (Ours)                  & 69.16\%    &  69.35\%                        &72.13\%& 70.64\% & 63.60\%    &  54.56\%                        &56.65\% &55.49\%\\
LRP4RAG$_\textrm{LSTM}$ (Ours)                  &  58.75\%   &       58.64\%                   & 72.83\% & 63.76\% &  69.87\%   &      68.26\%                   & 47.19\% & 55.56\%\\

LRP4RAG$_\textrm{LLM}$ (Ours)                  & \textbf{73.41\%}   &      \textbf{71.18\%}                   & 75.78\% & \textbf{73.54\%} &  \textbf{71.70\%}   &       \textbf{77.14\%}                   & 74.58\% & \textbf{75.86}\%\\

\bottomrule
\end{tabular}}
\caption{Comparison of LRP4RAG and baselines on the RAGTruth dataset. RAGTruth$_\textrm{Llama-7B}$ refers to using Llama-7B as generator, RAGTruth$_\textrm{Llama-13B}$ refers to using Llama-13B as generator.}\label{tab:main-results}
\end{table*}

The comparison results on Dolly-15k are listed in Table~\ref{tab:dolly}.  For threshold-based baselines, their optimal thresholds are  provided. On Dolly-15k$_\textrm{Qwen2.5-3B}$, LRP4RAG$_\textrm{LLM}$ outperforms the second-best SEP by 4.1\% in accuracy, 2.61\% in precision, 3.72\% in recall, and 3.15\% in F1-score. Additionally, the PCC reaches 0.5233, which is 0.0399 higher than that of SEP, indicating a strong correlation between the consistency between pruned contexts and answer after context pruning and RAG hallucinations. On Dolly-15k$_\textrm{Qwen2.5-7B}$, LRP4RAG$_\textrm{LLM}$ outperforms SEP by 5.1\% in accuracy, 3.24\% in precision, 4.61\% in recall, 3.83\% in F1-score and 0.025 in PCC. Compared to Dolly-15k$_\textrm{Qwen2.5-3B}$, the performance of LRP4RAG$_\textrm{LLM}$ on Dolly-15k$_\textrm{Qwen2.5-7B}$ slightly decreases, with accuracy decreasing by 1\%, precision by 0.95\%, recall by 0.71\%, F1-score by 0.91\%, and PCC by 0.0466. We find that Qwen-7B generates fewer hallucinated  samples compared to Qwen-3B, and the remaining hallucinated samples are more difficult to detect. Additionally, the answers in these hallucinated  samples are associated with multiple parts of the context, requiring a comprehensive analysis of longer contexts to derive the correct answer. As a result, the model tends to return longer contexts when extracting relevant information, which weakens the effectiveness of context pruning and increases the difficulty of consistency check.

\begin{table*}[htbp]
\centering

\resizebox{\textwidth}{!}{
\begin{tabular}{c|l|c|cccccc}
\toprule
& \textbf{Methods} & \textbf{Threshold} & \textbf{Accuracy} & \textbf{Precision} & \textbf{Recall} & \textbf{F1} & \textbf{AUC} & \textbf{PCC} \\
\midrule
\multirow{8}{*}{\textbf{Dolly$_\textrm{Qwen2.5-3B}$}} & Perplexity & -0.2213 & 56.50\% & 69.08\% & 56.52\% & 62.17\% & 60.02\% & 0.3087 \\
 & Energy & 26.6272 & 57.15\% & 70.31\% & 55.81\% & 62.23\% & 59.69\% & -0.0848 \\
 & LN-Entropy & -0.1362 & 55.35\% & 66.84\% & 58.33\% & 62.30\% & 55.23\% & 0.2347 \\
 & LexicalSim & 0.3906 & 60.90\% & 70.31\% & 66.08\% & 68.13\% & 61.76\% & 0.4103 \\
 & EigenScore & 1.1346 & 59.15\% & 68.88\% & 64.58\% & 66.66\% & 59.71\% & 0.2901 \\
 & SAPLMA & - & 70.70\% & 74.46\% & 78.17\% & 76.27\% & 73.25\% & 0.4587 \\
 & SEP & - & 73.10\% & 77.94\% & 79.19\% & 78.56\% & 76.58\% & 0.4832 \\
  & SelfCheckGPT & - & 53.15\% & 67.32\% & 32.47\% & 43.88\% & - & - \\
  & Prompt & - & 41.30\% & 58.41\% & 24.98\% & 34.99\% & - & - \\
  & Finetune & - & - & - & - & - & - & - \\
   & LRP4RAG$_\textrm{RF}$ (Ours) & - & 61.05\% & 63.91\% & 88.59\% & 74.11\% & - & - \\
  & LRP4RAG$_\textrm{MLP}$ (Ours) & - & 62.55\% & 66.99\% & 80.28\% & 73.05\% & - & - \\
 & LRP4RAG$_\textrm{SVM}$ (Ours) & - & 64.20\% & 66.12\% & 88.94\% & 75.84\% & - & - \\
  & LRP4RAG$_\textrm{LSTM}$ (Ours) & - & 58.50\% & 69.60\% & 61.96\% & 65.56\% & - & - \\
 & LRP4RAG$_\textrm{LLM}$ (Ours) & - & \textbf{77.20\%} & \textbf{80.55\%} & \textbf{82.91\%} & \textbf{81.71\%} & - & \textbf{0.5233} \\
\cmidrule(lr){1-9}
\cmidrule(lr){1-9}
\multirow{8}{*}{\textbf{Dolly$_\textrm{Qwen2.5-7B}$}} & Perplexity & -0.2431 & 48.55\% & 51.14\% & 61.15\% & 55.72\% & 53.72\% & 0.1729 \\
 & Energy & 26.3421 & 57.65\% & 63.71\% & 67.50\% & 65.55\% & 62.49\% & -0.0768 \\
 & LN-Entropy & -0.1487 & 55.05\% & 60.47\% & 65.72\% & 62.98\% & 60.47\% & 0.2748 \\
 & LexicalSim & 0.4232 & 63.40\% & 62.21\% & 75.60\% & 68.25\% & 68.43\% & 0.4374 \\
 & EigenScore & 1.0823 & 57.95\% & 58.57\% & 70.03\% & 63.79\% & 64.78\% & 0.3212 \\
 & SAPLMA & - & 69.10\% & 74.62\% & 76.06\% & 75.33\% & 72.53\% & 0.4493 \\
 & SEP & - & 71.10\% & 76.36\% & 77.59\% & 76.97\% & 74.68\% & 0.4517 \\
 & SelfCheckGPT & - & 53.15\% & 67.32\% & 32.47\% & 43.88\% & - & - \\
  & Prompt & - & 47.95\% & 61.46\% & 47.23\% & 53.36\% & - & - \\
  & Finetune & - & - & - & - & - & - & - \\
    & LRP4RAG$_\textrm{RF}$ (Ours) & - & 62.75\% & 64.17\% & 93.11\% & 75.93\% & - & - \\
  & LRP4RAG$_\textrm{MLP}$ (Ours) & - & 59.20\% & 66.88\% & 70.19\% & 68.48\% & - & - \\
 & LRP4RAG$_\textrm{SVM}$ (Ours) & - & 63.60\% & 65.83\% & 88.23\% & 75.39\% & - & - \\
  & LRP4RAG$_\textrm{LSTM}$ (Ours) & - & 61.70\% & 65.67\% & 89.80\% & 73.75\% & - & - \\
 & LRP4RAG$_\textrm{LLM}$ (Ours) & - & \textbf{76.20\%} & \textbf{79.60\%} & \textbf{82.20\%} & \textbf{80.80\%} & - & \textbf{0.4767} \\
\bottomrule
\end{tabular}}
\caption{Comparison of LRP4RAG and baselines on the Dolly-15k dataset. Dolly-15k$_\textrm{Qwen2.5-3B}$ refers to using Qwen2.5-3B as generator, Dolly-15k$_\textrm{Qwen2.5-7B}$ refers to using Qwen2.5-7B as generator.}
\label{tab:dolly}
\end{table*}

\subsection{RQ3: LRP4RAG$_\textrm{Classifier}$ vs LRP4RAG$_\textrm{LLM}$}
\textbf{Experimental Design.} In RQ3, we aim to compare the differences between the two variants of LRP4RAG in terms of effectiveness, computational complexity, and robustness. 

\textbf{Results. }
Table~\ref{tab:main-results} and Table~\ref{tab:dolly} present the performance comparison between LRP4RAG$_\textrm{Classifier}$ and LRP4RAG$_\textrm{LLM}$. We analyze effectiveness, computational complexity, and robustness as follows:

\textbf{Effectiveness Analysis.} LRP4RAG$_\textrm{LLM}$ consistently outperforms LRP4RAG$_\textrm{Classifier}$ on the RAGTruth dataset. Specifically, on RAGTruth$_\textrm{Llama-7B}$, LRP4RAG$_\textrm{LLM}$ achieves 73.4\% accuracy and an F1-score of 73.5\%, surpassing the best-performing LRP4RAG$_\textrm{Classifier}$ (LRP4RAG$_\textrm{SVM}$), which attains 69.2\% accuracy and an F1-score of 70.6\%. A similar trend is observed on RAGTruth$_\textrm{Llama-13B}$, where LRP4RAG$_\textrm{Classifier}$ exhibits higher recall but suffers from a significant drop in precision.
Similarly, on the Dolly-15k dataset, LRP4RAG$_\textrm{LLM}$ achieves state-of-the-art results. 
% On {\textbf{Dolly-15k$_\textrm{Qwen2.5-3B}$}}, it gains 77.2\% accuracy, 80.5\% precision, 82.9\% recall, and an 81.7\% F1-score. With {\textbf{Dolly-15k$_\textrm{Qwen2.5-7B}$}}, it achieves 76.2\% accuracy, 79.6\% precision, 82.2\% recall, and an 80.8\% F1-score.
The advantage of LRP4RAG$_\textrm{LLM}$ lies in its ability to leverage the reasoning and contextual comprehension capabilities of LLMs. This enables direct and precise consistency detection without additional training, effectively capturing subtle contextual nuances and achieving superior performance.
In contrast, LRP4RAG$_\textrm{Classifier}$ often suffers from lower accuracy and precision, as tuning for high recall can lead to false positives. Moreover, its performance varies across different dataset, necessitating frequent re-tuning.

\begin{figure}[htbp]
\centering
\includegraphics[width=0.5\textwidth]{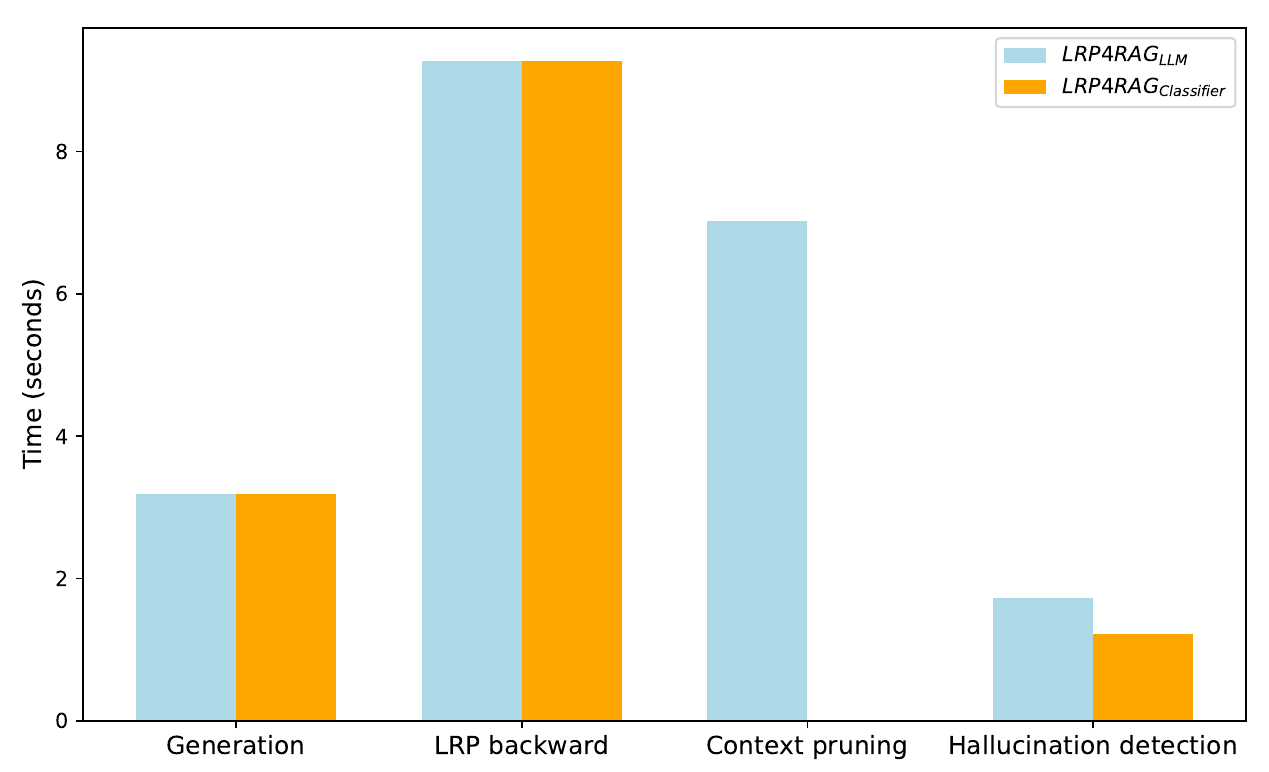} % Replace with your image path
        \caption{Comparison of time cost between LRP4RAG$_\textrm{Classifier}$ and LRP4RAG$_\textrm{LLM}$ on Dolly-15k using Qwen2.5-3B.}
        \label{fig:time-comparison}
\end{figure}

\textbf{Computational Complexity Analysis.} However, despite its superior accuracy, LRP4RAG$_\textrm{LLM}$ requires significantly greater computational resources and longer processing times due to its reliance on large-scale model inference. 
This may pose deployment challenges in resource-limited or latency-sensitive environments. 
In contrast, LRP4RAG$_\textrm{Classifier}$ provides computational efficiency, ease of deployment, and greater interpretability through explicit feature analysis, along with flexibility in tuning to achieve desired recall levels. As shown in Figure~\ref{fig:time-comparison}, we present the average time taken for hallucination detection of a single sample by the two methods. LRP4RAG$_\textrm{LLM}$ takes an average of 21.21 seconds per sample, while LRP4RAG$_\textrm{Classifier}$ takes only 13.68 seconds. LRP4RAG$_\textrm{LLM}$ consumes 55.07\% more time than LRP4RAG$_\textrm{Classifier}$. Specifically, the LRP backward stage is the most time-consuming part of LRP4RAG. The speed of LRP backward is 0.131 seconds per token, which is approximately three times that of generation stage at 0.045 seconds per token. The performance loss of LRP4RAG$_\textrm{LLM}$ compared to LRP4RAG$_\textrm{Classifier}$ is mainly concentrated in the context pruning stage, which takes an average of about 7 seconds. In contrast, LRP4RAG$_\textrm{Classifier}$ directly uses a classifier for hallucination detection, making it more efficient overall.

% 需要增加点时间对比

\textbf{Robustness Analysis.}
LRP4RAG$_\textrm{LLM}$ demonstrates stronger overall robustness in performance compared to LRP4RAG$_\textrm{Classifier}$. On the RAGTruth dataset, detecting hallucinations in outputs from RAGTruth$_\textrm{Llama-13B}$ proves more challenging for all methods than in RAGTruth$_\textrm{Llama-7B}$. Both baselines and LRP4RAG$_\textrm{Classifier}$ experience a drop in recall when applied to RAGTruth$_\textrm{Llama-7B}$.
For example, the recall of LRP4RAG$_\textrm{Classifier}$ drops from 72\% on RAGTruth$_\textrm{Llama-7B}$ to 47\%-57\% on RAGTruth$_\textrm{Llama-13B}$, indicating that it misses more true answers from the larger model. In contrast, LRP4RAG$_\textrm{LLM}$ remains effective on RAGTruth$_\textrm{Llama-13B}$. This suggests that hallucinations from RAGTruth$_\textrm{Llama-13B}$, which may be more contextually entangled, can still be detected by the consistency check of LRP4RAG$_\textrm{LLM}$, whereas LRP4RAG$_\textrm{Classifier}$ struggles to generalize as well to the larger model’s outputs.
We observe a similar pattern on Dolly$_\textrm{Qwen2.5-3B}$ and Dolly$_\textrm{Qwen2.5-7B}$. Upgrading from Dolly$_\textrm{Qwen2.5-3B}$ to Dolly$_\textrm{Qwen2.5-7B}$ results in only a slight decrease in the performance of LRP4RAG$_\textrm{LLM}$.
LRP4RAG$_\textrm{LLM}$ demonstrates consistent robustness, whereas LRP4RAG$_\textrm{Classifier}$ is less stable and often requires calibration for each model to maintain performance.

\subsection{RQ4: Ablation Study}
\textbf{Experimental Design.} In LRP4RAG$_\textrm{LLM}$, hallucination detection is evaluated by analyzing the consistency between the pruned contexts ($\textbf{\textit{C}}_{\text{key}}$, $\textbf{\textit{C}}_{\text{llm}}$) and the model output. To investigate the contribution of each of the three consistency measures defined in Equations~\ref{eq-20}–\ref{eq-22}, we perform an ablation study using Qwen-3B on the Dolly-15k dataset, building upon the original experimental setup. For each consistency measure, we remove it from the prompt and instruct the LLM to assess the remaining two. The results are then compared with those obtained under the full consistency setting to evaluate the effectiveness of each individual consistency in hallucination detection.

\textbf{Consistency between $\textbf{\textit{C}}_{\text{key}}$ and $\textbf{\textit{C}}_{\text{llm}}$ ($\textbf{\textit{Consistency}}_{\text{1}}$). } 
$Consistency_1$ reflects whether the model's internal thoughts match its words. Inconsistency of the pruned contexts indicates high uncertainty which is a cause of hallucination. As shown in the second row of Table~\ref{tab:ablation}, removing the check for $Consistency_1$ leads to a decrease in accuracy, recall, and F1-score by 3.4\%, 3.78\%, and 2.01\%, respectively, demonstrating the importance and efficiency of $Consistency_1$.

\begin{figure*}[htbp]
\centering
\includegraphics[width=\textwidth]{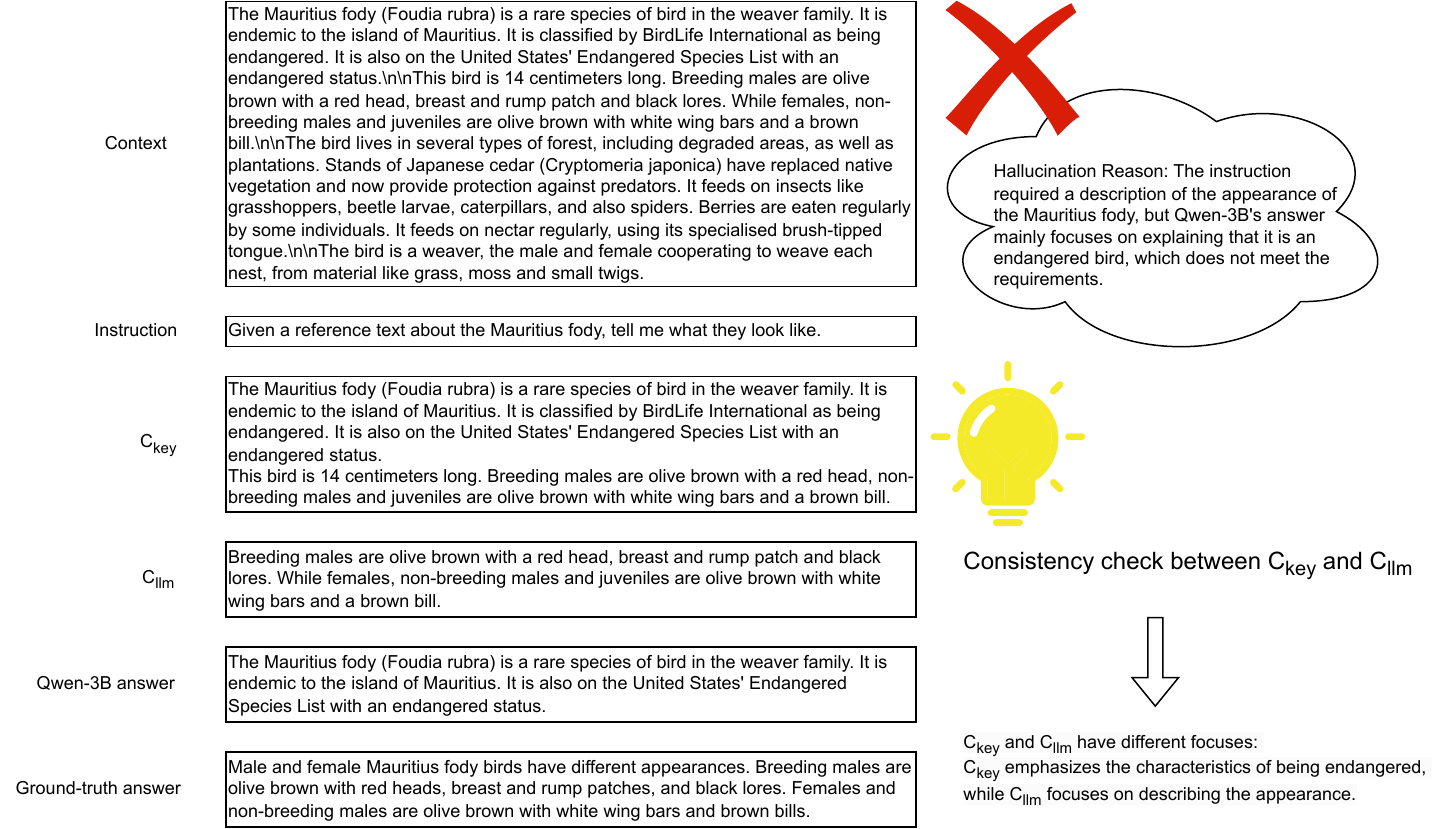} % Replace with your image path
        \caption{Example to demonstrate the effectiveness of $Consistency_1$. When removing the check for $Consistency_1$, Qwen-3B fails to identify the RAG hallucination shown in the picture.}
        \label{fig:ablation-demo1}
\end{figure*}

As shown in Figure~\ref{fig:ablation-demo1}, when we remove the check for $Consistency_1$, Qwen-3B assumes there is no hallucination, which is a mistaken judgment. The ground-truth answer focuses on describing the appearance of the Mauritius fody, while the answer emphasizes that the Mauritius fody is an endangered bird, not aligning with the instruction. When introducing the check for $Consistency_1$, Qwen-3B compares the two pruned contexts and easily identifies their inconsistency: $\textit{C}_{\text{key}}$ emphasizes that the Mauritius fody is an endangered bird and only mentions its appearance in the last sentence, while $\textit{C}_{\text{llm}}$ details the differences in appearance between male and female Mauritius fodies. Based on such observation, Qwen-3B correctly identifies the hallucination. Notably, the pruned contexts are much shorter than the full context, helping Qwen-3B to fully understand.

\textbf{Consistency between $\textbf{\textit{C}}_{\text{key}}$ and answer ($\textbf{\textit{Consistency}}_{\text{2}}$). } $Consistency_2$ reflects whether the model correctly expresses its internal thoughts. Even though the model may sometimes understand the context correctly, the output answer is still not accurate enough. After pruning the context based on relevance, a shorter context allows the model to more easily capture the differences between its output and its internal thoughts. This approach is similar to a self-check, but the results show that consistency checks based on context pruning are much more effective than direct hallucination detection by prompting the model. Using the Prompt method, Qwen-3B achieves only a 41.30\% accuracy in hallucination detection, while LRP4RAG$_\textrm{LLM}$ exceeds 70\% accuracy. To further highlight the importance of $Consistency_2$, we remove it from LRP4RAG$_\textrm{LLM}$. The results show that without $Consistency_2$, the accuracy of LRP4RAG$_\textrm{LLM}$ decreases by 5.70\%, recall decreases by 8.45\%, and F1-score decreases by 3.06\%, demonstrating the effectiveness of $Consistency_2$.

\textbf{Consistency between $\textbf{\textit{C}}_{\text{llm}}$ and answer ($\textbf{\textit{Consistency}}_{\text{3}}$). } $Consistency_3$ reflects whether the model can maintain consistency across relevant tasks. We design a new task where the model is required to find relevant context based on the question, which is strongly related to the original RAG task. The context output by the model should be consistent with the answer generated by RAG. Any contradictions indicate the occurrence of hallucinations. To validate the effectiveness of $Consistency_3$, we remove it from LRP4RAG$_\textrm{LLM}$ and compare with the original results. We find that accuracy decreases by 7\%, recall decreases by 9.1\%, and F1-score decreases by 4.17\%.

\begin{table*}[htbp]
\centering

\resizebox{0.8\textwidth}{!}{
\begin{tabular}{l|cccc}
\toprule
\textbf{Consistency}  & \textbf{Accuracy} & \textbf{Precision} & \textbf{Recall} & \textbf{F1} \\
\midrule
$Consistency_{1} + Consistency_{2} + Consistency_{3}$& 77.20\% & 80.55\% & 82.91\% & 81.71\% \\
$ Consistency_{2} + Consistency_{3}$& 73.80\% & 81.34\% & 78.13\% & 79.70\%\\
$ Consistency_{1} + Consistency_{3}$& 71.50\% & 83.33\% & 74.46\% & 78.65\%\\
$ Consistency_{1} + Consistency_{2}$& 70.20\% & 81.66\% & 73.81\% & 77.54\%\\
\bottomrule
\end{tabular}}
\caption{The effeciency of three types of consistency check on LRP4RAG$_\textrm{LLM}$ (An ablation study conducted on Dolly-15k, using Qwen-3B). }
\label{tab:ablation}
\end{table*}

\subsection{RQ5: The Impact of Hyperparameters on LRP4RAG}
\textbf{Experimental Design.}
In RQ5, we investigate the impact of two important hyperparameters on the performance of LRP4RAG. The first hyperparameter is length of feature vectors used in LRP4RAG$_\textrm{Classifier}$, the second hyperparameter is the portion of pruned context used in LRP4RAG$_\textrm{LLM}$.

\textbf{Vector length.} For LRP4RAG$_\textrm{Classifier}$, we investigate how the length of feature vectors influence hallucination classification performance. The relevance matrix is preprocessed (e.g., average pooling) into feature vectors, and the length of vectors determines the completeness of the information retained. Excessively long vectors can introduce additional noise, while vectors with insufficient length can cause information loss, leading to a degradation in hallucination detection performance. In search for the most effective vector length $L^{*}$ , we test with different $L^{feature}$ in the interval (0,500). In Table \ref{tab:resampling}, we present the effectiveness of LRP4RAG$_\textrm{Classifier}$ on the RAGTruth$_\textrm{Llama-13B}$ dataset applying different $L^{feature}$. We find that when $L^{feature}$ is small, the performance of LRP4RAG$_\textrm{Classifier}$ improves rapidly as $L^{feature}$
increases. However, after exceeding a certain threshold (220), the performance of LRP4RAG$_\textrm{Classifier}$ begins to decline slowly. Check Appendix~\ref{full-resampling} for full results.

\begin{table}[htbp]
\centering

\begin{tabular}{l|ccc} 
\toprule
$L^{feature}$ & \textbf{Accuracy} & \textbf{Precision} & \textbf{Recall}   \\

\midrule
% $L^{feature}$=100                  & 67.13\%    &  66.57\%                        &\textbf{72.96\%}  \\

% $L^{feature}$=200                  & 67.24\%    &  67.56\%                        &70.21\%  \\

$L^{feature}$ = 205                  & 66.63\%    &  66.70\%                        &70.16\%  \\

$L^{feature}$ = 210                  & 66.43\%    &  67.75\%                        &67.24\%  \\

$L^{feature}$ = 215                  & 66.13\%    &  66.48\%                        &69.51\%  \\

$L^{feature}$ = 220                  & \textbf{69.16\%}    &  69.35\%                        &\textbf{72.13\%}  \\

$L^{feature}$ = 225                  & 68.25\%    &  68.52\%                        &70.84\%  \\

$L^{feature}$ = 230                  & 68.65\%    &  \textbf{69.76\%}                        &69.36\%  \\

$L^{feature}$ = 235                  & 65.93\%    &  66.55\%                        &68.02\%  \\

% $L^{feature}$=240                  & 67.84\%    &  69.12\%                        &67.91\%  \\

% $L^{feature}$=300                  & 64.71\%    &  65.96\%                        &65.03\%  \\

\bottomrule
\end{tabular}

\caption{The performance of LRP4RAG$_\textrm{Classifier}$ under different $L^{feature}$ on the RAGTruth$_\textrm{Llama-7B}$ dataset.}\label{tab:resampling}

\end{table}

\begin{figure}[htbp]
\centering
\includegraphics[width=0.5\textwidth]{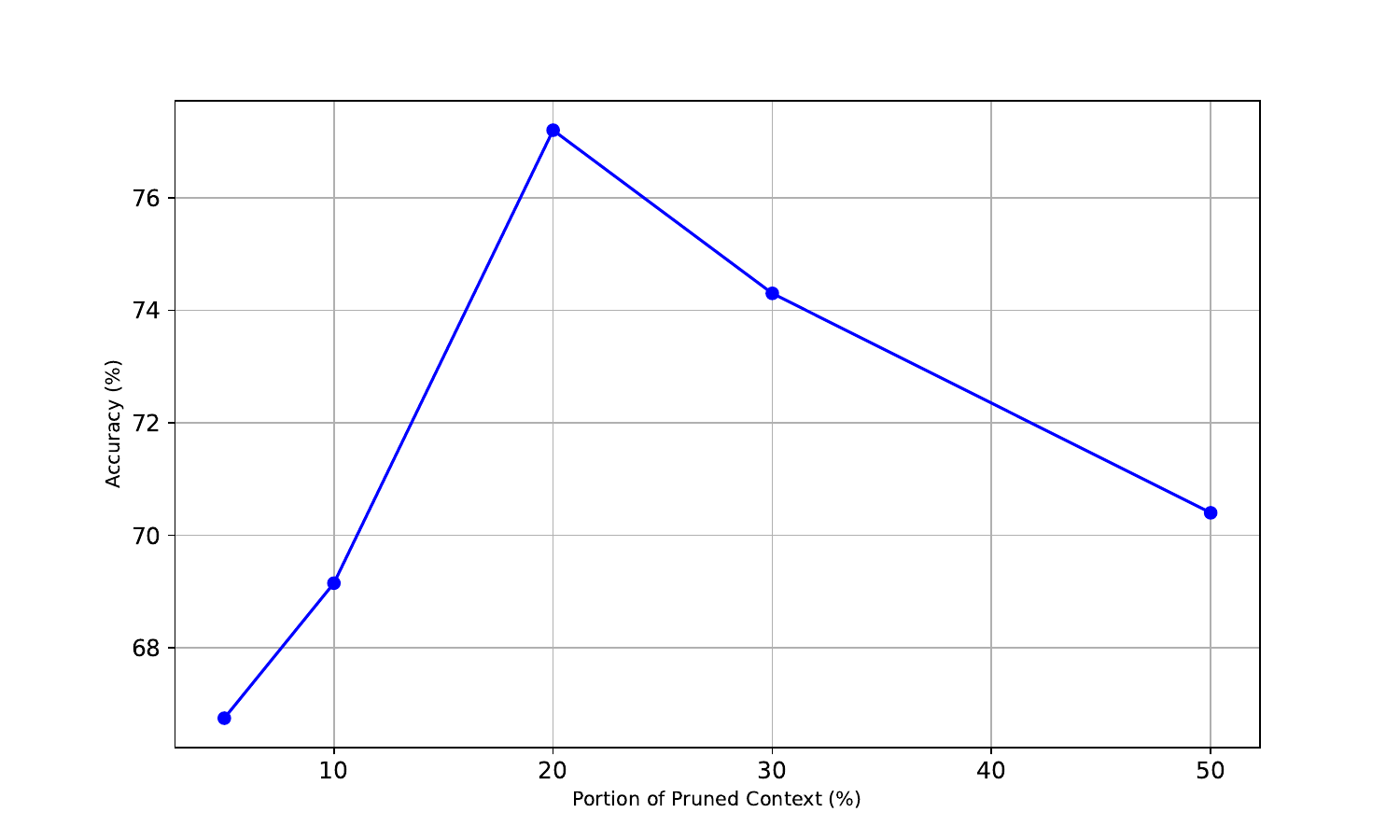} % Replace with your image path
        \caption{The relationship between hallucination detection accuracy and the proportion of context pruning.}
        \label{fig:portion}
\end{figure}

\textbf{Portion of pruned context.} For LRP4RAG$_\textrm{LLM}$, we select context segments that are highly relevant to the answer according to sentence-level relevance $R^{sentence}$, which is referred to as context pruning. Through context pruning, we retain the most important parts of the context while removing redundancies, which helps the model to more directly and efficiently self-verify the correctness of the answer. We find that the performance of pruning is closely related to the proportion of context retained. Insufficient context may fail to provide enough cues for reliable inference, while overly long context can introduce noise and complicate the identification of relevant information.  We adjust the value of $K$ in Equation~\ref{eq-19} to control the length of the context after pruning. As shown in Figure~\ref{fig:portion}, we test LRP4RAG$_\textrm{LLM}$ for $K$ = 5, 10, 20, 30, and 50. The accuracy is lowest when $K$ = 5, at only 66.75\%, and highest when $K$ = 20, at 77.2\%. We find that when $K$ is too small, the hallucination detection accuracy of LRP4RAG$_\textrm{LLM}$ significantly decreases due to the loss of too much critical information. Conversely, when $K$ is large, the retained context becomes too redundant, leading to a certain decrease in the performance of LRP4RAG$_\textrm{LLM}$.

\section{Related Works}
\subsection{LLM Hallucination}\label{related-works-hallucination}

LLM Hallucination derives from inconsistency with facts or context~\citep{xu2025hallucinationinevitableinnatelimitation}. Existing methods for detecting hallucinations are typically categorized into four types, detailed as follows. 

\begin{itemize}
    \item \textbf{Task-specific methods.} Task-specific methods primarily concentrate on
detecting hallucinations in specific natural language generation tasks, such as machine translation~\citep{DBLP:conf/acl/DaleVBC23,DBLP:conf/eacl/GuerreiroVM23}, dialogue generation~\citep{dziri-etal-2022-evaluating} and question answering~\citep{durmus-etal-2020-feqa}. Ma et al.~\citep{JIN2025107291} propose a graph-based approach for hallucination detection. CSL-VQA~\citep{CAO2025111129} combines self-supervised and contrastive learning to detect and mitigate hallucinations in Visual Question Answering. Causal-ViT~\citep{LI2023107123} introduces an approximate intervention strategy to mitigate hallucinations in visual models while improving robustness. Other works also explore hallucination detection in the field of AI4Science~\citep{he2025hybrid,guo2025assessing,he2024comparative,guo2024performance,guo2024monthly,guo2023prediction}, such as climate prediction and temperature forecasting.
    \item \textbf{Consistency-based methods.} Consistency-based methods~\citep{DBLP:conf/iclr/0002WSLCNCZ23} apply slight perturbations to LLM input and then evaluate the self-consistency of the outputs. If there exists significant divergence among answers, it suggests hallucinations happen.   Similar methods~\citep{shi-etal-2022-natural} prompt LLMs to generate multiple responses to the same question and evaluating the self-consistency of those responses.
    \item \textbf{Uncertainty-based methods.} Token-level uncertainty~\citep{DBLP:conf/iclr/MalininG21,huang2023lookleapexploratorystudy} estimation (e.g. entropy) has proven efficient in hallucination detecting on NLP tasks. Recent studies~\citep{duan2024shiftingattentionrelevancepredictive,yin2023largelanguagemodelsknow} further work on sentence-level and language-level uncertainty to make extensive progress.
    \item \textbf{LLM-based methods.} LLM-based methods~\citep{arteaga2024hallucinationdetectionllmsfast,bhamidipati2024zeroshotmultitaskhallucinationdetection} design prompts or directly fine-tune LLMs to detect hallucination.
\end{itemize}     

Despite the extensive research on LLM hallucination, yet RAG hallucination receives few attention. Inspired by the study~\citep{DBLP:journals/tacl/LiuLHPBPL24} on long contexts, as well as the theory of additive interpretability~\citep{agarwal2021neuraladditivemodelsinterpretable,10.1007/978-3-030-28954-6_10}, we first analyze the causes of RAG hallucination and then apply context pruning to check the consistency between model's internal states and outputs. The approach LRP4RAG we propose is the first to apply transformer-based LRP to RAG hallucination, leading to more accurate results than previous approaches.

\subsection{Retrieval-Augmented Generation}
Traditional NLP tasks have been benefiting from RAG since ChatGPT appears. RAG enhances LLMs by retrieving relevant document chunks from outside knowledege base.  By referencing external knowledge, RAG effectively
reduces the problem of LLM hallucination. However, recent studies show RAG still faces problems which make hallucination inevitable. Barnett~\citep{barnett2024sevenfailurepointsengineering} lists 7 main problems identified from case studies, including insufficient context, missing top ranked documents, etc. These problems lead to a degradation in RAG output quality, ultimately resulting in hallucinations. The work "Lost in the Middle"~\citep{DBLP:journals/tacl/LiuLHPBPL24} also emphasizes the difficulty that LLMs encounter when attempting to pinpoint the most pertinent information within extensive contextual spans. This deficiency in handling long contextual information can lead to hallucinations when RAG generates text. Recognizing the numerous limitations of RAG, we delve into an exhaustive analysis of the distinct internal states of LLMs that accompany the occurrence of hallucinations. Capitalizing on these insights, we propose a novel methodology, LRP4RAG, designed to detect hallucinations during the generation process with RAG.

\section{THREATS TO VALIDITY}
\textbf{Internal Threat.} The main internal threat is the potential data leakage in LRP4RAG. LRP4RAG$_\textrm{LLM}$ is implemented with Qwen, and the pre-training data of Qwen may overlap with the benchmarks used in our study. To address this concern, we carefully inspect the pre-training data of Qwen, especially the conversational and book-style corpora. We find that there is no overlap between these datasets and the two benchmarks (e.g., Dolly-15k and RAGTruth) used in our study. It is worth noting that the data leakage concern motivates our choice of open-source LLMs, instead of more powerful black-box LLMs. Thus, we confidently ensure that the pre-training data does not contain any overlap with the evaluation datasets in our experiments. 

\textbf{External Threat.} The main external threat to validity lies in the use of QA-type benchmarks to evaluate LRP4RAG. As a result, the performance of LRP4RAG may not extended to other types of RAG datasets. To address this issue, we retain a portion of non-QA data from Dolly-15k (e.g., classification, information extraction, and brainstorming tasks) to ensure the broad applicability of LRP4RAG across various types of RAG datasets.
Moreover, the two high-quality benchmark datasets adopted in our experiments are representative within the hallucination detection field, and all baselines also utilize these datasets to draw reliable conclusions. Therefore, we believe that the impact of this threat on our findings is relatively limited. In the future, we will further explore the performance of LRP4RAG on new RAG benchmarks of different types.

\textbf{Construct Threat.} The main construct threat to validity comes from the evaluation metrics. Previous studies~\citep{lin-etal-2022-towards,malinin2021uncertaintyestimationautoregressivestructured,ren2023outofdistributiondetectionselectivegeneration} often neglect accuracy and precision, focusing primarily on recall and F1. Although a high recall indicates that most hallucinated samples can be detected, low precision and accuracy may still lead to the misclassification of normal samples as hallucinations, which negatively affects the overall performance of the method.
To more comprehensively assess the overall performance of LRP4RAG and other baselines, we consider all four metrics, including accuracy, precision, recall, and F1. This allows for a fairer and more balanced comparison among different approaches.

\section{Conclusion}
In this work, we propose an LRP-based approach LRP4RAG to detect RAG hallucinations. 
Different from existing works that utilize RAG to mitigate LLM hallucination, we aim to discuss and explore vulnerability in RAG. 
We look into internal states of LLMs and consider the relevance between the context and answer. 
We find the difference of relevance distribution between normal and hallucinated samples, and use it as evidence for classification. 
Extensive experiments show LRP4RAG achieves performance gains over strong baselines.

\bibliographystyle{cas-model2-names}

\bibliography{cas-refs}

\appendix
\section{Full Threshold-based Results}
\label{full-threshold}

In Table \ref{tab:threshold-appendix}, we present the full threshold-based results. On RAGTruth$_\textrm{Llama-7B}$ dataset, we search for the optimal $\overline{R^{token}}$ within the interval (0.2, 0.4) and (0.4, 0.6) respectively, setting step size to 0.01. On RAGTruth$_\textrm{Llama-13B}$ dataset, we search for the optimal $\overline{R^{token}}$ within the interval (0.15, 0.35) and (0.3, 0.5) respectively, setting step size to 0.01.

\begin{table*}[htbp]
\centering

\resizebox{\textwidth}{!}{
\begin{tabular}{l|cccc|cccc}
\toprule
& \multicolumn{4}{c|}{RAGTruth$_\textrm{Llama-7B}$} & \multicolumn{4}{c}{RAGTruth$_\textrm{Llama-13B}$} \\

\midrule
Relevance Threshold &   Accuracy & Precision & Recall & F1 & Accuracy & Precision & Recall & F1  \\
\midrule

$\overline{R^{token}}$                     &         &           &        &         &           &       & & \\ 
\midrule
$t = 0.3$ & \textasciitilde & \textasciitilde & \textasciitilde & \textasciitilde & 58.85\% & 21.67\% & 1.25\% & 2.36\% \\
$t = 0.31$ & \textasciitilde & \textasciitilde & \textasciitilde & \textasciitilde & 58.95\% & 25.67\% & 1.47\% & 2.78\% \\
$t = 0.32$ & \textasciitilde & \textasciitilde & \textasciitilde & \textasciitilde & 58.95\% & 31.24\% & 2.19\% & 4.08\% \\
$t = 0.33$ & \textasciitilde & \textasciitilde & \textasciitilde & \textasciitilde & 59.25\% & 43.02\% & 3.7\% & 6.8\% \\
$t = 0.34$ & \textasciitilde & \textasciitilde & \textasciitilde & \textasciitilde & 59.05\% & 42.17\% & 4.68\% & 8.4\% \\
$t = 0.35$ & \textasciitilde & \textasciitilde & \textasciitilde & \textasciitilde & 59.66\% & 49.76\% & 6.45\% & 11.39\% \\
$t = 0.36$ & \textasciitilde & \textasciitilde & \textasciitilde & \textasciitilde & 59.76\% & 51.57\% & 8.22\% & 14.14\% \\
$t = 0.37$ & \textasciitilde & \textasciitilde & \textasciitilde & \textasciitilde & 59.35\% & 48.72\% & 10.31\% & 16.99\% \\
$t = 0.38$ & \textasciitilde & \textasciitilde & \textasciitilde & \textasciitilde & 59.35\% & 48.71\% & 12.53\% & 19.92\% \\
$t = 0.39$ & \textasciitilde & \textasciitilde & \textasciitilde & \textasciitilde & 59.45\% & 49.41\% & 15.29\% & 23.27\% \\
$t = 0.4$ & 54.9\% & 69.83\% & 21.85\% &33.16\% & 59.05\% & 47.77\% & 17.53\% & 25.6\% \\
$t = 0.41$ & 55.61\% & 68.33\% & 26.03\% &37.65\% & 59.36\% & 49.2\% & 23.54\% & 31.74\% \\
$t = 0.42$ & 57.23\% & 68.42\% & 31.96\% &43.53\% & 60.06\% & 50.96\% & 31.11\% & 38.44\% \\
$t = 0.43$ & 58.85\% & 67.39\% & 39.16\% &49.5\% & 60.37\% & 51.38\% & 36.12\% & 42.22\% \\
$t = 0.44$ & 58.34\% & 63.73\% & 44.25\% &52.22\% & 58.44\% & 48.29\% & 40.48\% & 43.86\% \\
$t = 0.45$ & 59.35\% & 63.07\% & 50.69\% &56.17\% & 57.94\% & 47.86\% & 47.85\% & 47.7\% \\
$t = 0.46$ & 60.36\% & 62.6\% & 56.54\% &59.36\% & 56.72\% & 46.87\% & 54.47\% & 50.23\% \\
$t = 0.47$ & 60.16\% & 61.0\% & 61.85\% &61.38\% & 55.81\% & 46.3\% & 60.26\% & 52.22\% \\
$t = 0.48$ & 60.26\% & 59.72\% & 69.49\% &64.23\% & 54.7\% & 45.78\% & 67.2\% & 54.33\% \\
$t = 0.49$ & 59.86\% & 58.49\% & 75.21\% &65.77\% & 53.48\% & 45.27\% & 73.92\% & 56.04\% \\
$t = 0.5$ & 60.16\% & 58.06\% & 81.42\% &67.72\% & 51.87\% & 44.56\% & 79.09\% & 56.87\% \\
$t = 0.51$ & 59.76\% & 57.35\% & 85.34\% &68.55\% & \textasciitilde & \textasciitilde & \textasciitilde & \textasciitilde  \\
$t = 0.52$ & 59.15\% & 56.55\% & 90.06\% &69.42\% & \textasciitilde & \textasciitilde & \textasciitilde & \textasciitilde  \\
$t = 0.53$ & 58.24\% & 55.74\% & 92.58\% &69.53\% & \textasciitilde & \textasciitilde & \textasciitilde & \textasciitilde  \\
$t = 0.54$ & 57.53\% & 55.12\% & 94.88\% &69.69\% & \textasciitilde & \textasciitilde & \textasciitilde & \textasciitilde  \\
$t = 0.55$ & 56.02\% & 54.11\% & 97.05\% &69.43\% & \textasciitilde & \textasciitilde & \textasciitilde & \textasciitilde  \\
$t = 0.56$ & 55.01\% & 53.47\% & 98.42\% &69.25\% & \textasciitilde & \textasciitilde & \textasciitilde & \textasciitilde  \\
$t = 0.57$ & 53.69\% & 52.73\% & 98.61\% &68.67\% & \textasciitilde & \textasciitilde & \textasciitilde & \textasciitilde  \\
$t = 0.58$ & 53.29\% & 52.5\% & 99.0\% &68.57\% & \textasciitilde & \textasciitilde & \textasciitilde & \textasciitilde  \\
$t = 0.59$ & 53.09\% & 52.38\% & 99.43\% &68.57\% & \textasciitilde & \textasciitilde & \textasciitilde & \textasciitilde  \\
$t = 0.6$ & 52.88\% & 52.27\% & 99.63\% &68.52\% & \textasciitilde & \textasciitilde & \textasciitilde & \textasciitilde  \\
\bottomrule
\end{tabular}}
\caption{Full results of the threshold-based method to validate the efficiency of relevance on RAG hallucination detection. }
\label{tab:threshold-appendix}
\end{table*}

\section{Full Results of Classifier-based LRP4RAG }
\label{full-resampling}

When testing the impact of feature vector length $L^{feature}$ on classification results, we set the step size to 10 in the interval (0, 150), to 5 in the interval (150, 300), and to 50 in the interval (300, 500). From the results in Table~\ref{tab:resampling-appendix}, it can be observed that the classifier's performance is not optimal when $L^{feature}$ is too large or too small. Especially when $L^{feature}$ is excessively small (less than 30), the feature vector loses too much information that the classifier tends to make extreme decisions, categorizing the majority of samples as hallucinations. Conversely, when $L^{feature}$ is excessively large (over 300), the classifier's output stabilizes; however, due to the presence of noise, it fails to achieve better outcomes. The findings indicate that the classifier achieves better results when L falls within the interval (220, 230), with $L^{feature}$ = 220 yielding the optimal outcome. Resampling under this condition effectively reduces noise in the processed data while preserving crucial features.

\begin{table*}[htbp]
\centering

\resizebox{0.45\textwidth}{!}{
\begin{tabular}{l|ccc} 
\toprule
$L^{feature}$ & Accuracy & Precision & Recall   \\

\midrule
$L^{feature}$=10                  & 55.92\%    &  54.58\%                        &88.42\%  \\

$L^{feature}$=20                  & 59.75\%    &  57.82\%                        &81.88\%  \\

$L^{feature}$=30                  & 61.98\%    &  60.11\%                        &77.82\%  \\

$L^{feature}$=40                  & 65.11\%    &  63.23\%                        &76.83\%  \\

$L^{feature}$=50                  & 66.63\%    &  65.04\%                        &75.80\%  \\

$L^{feature}$=60                  & 66.83\%    &  66.27\%                        &72.67\%  \\

$L^{feature}$=70                  & 66.73\%    &  66.44\%                        &72.96\%  \\

$L^{feature}$=80                  & 67.13\%    &  66.57\%                        &72.96\%  \\

$L^{feature}$=90                  & 68.55\%    &  67.69\%                        &74.82\%  \\

$L^{feature}$=100                  & 67.13\%    &  66.57\%                        &72.96\%  \\

$L^{feature}$=110                  & 67.34\%    &  67.77\%                        &70.28\%  \\

$L^{feature}$=120                  & 67.04\%    &  67.22\%                        &70.27\%  \\

$L^{feature}$=130                  & 67.54\%    &  67.67\%                        &71.12\%  \\

$L^{feature}$=140                  & 67.64\%    &  67.65\%                        &71.27\%  \\

$L^{feature}$=150                  & 67.44\%    &  67.94\%                        &70.16\%  \\

$L^{feature}$=155                  & 66.33\%    &  66.07\%                        &70.58\%  \\

$L^{feature}$=160                  & 66.63\%    &  67.20\%                        &68.46\%  \\

$L^{feature}$=165                  & 67.14\%    &  67.71\%                        &69.44\%  \\

$L^{feature}$=170                  & 66.33\%    &  66.24\%                        &71.13\%  \\

$L^{feature}$=175                  & 67.04\%    &  67.11\%                        &70.86\%  \\

$L^{feature}$=180                  & 67.74\%    &  68.17\%                        &70.26\%  \\

$L^{feature}$=185                  & 66.83\%    &  66.59\%                        &71.67\%  \\

$L^{feature}$=190                  & 66.23\%    &  66.02\%                        &71.08\%  \\

$L^{feature}$=195                  & 66.33\%    &  66.51\%                        &70.51\%  \\

$L^{feature}$=200                  & 67.24\%    &  67.56\%                        &70.21\%  \\

$L^{feature}$=205                  & 66.63\%    &  66.70\%                        &70.16\%  \\

$L^{feature}$=210                  & 66.43\%    &  67.75\%                        &67.24\%  \\

$L^{feature}$=215                  & 66.13\%    &  66.48\%                        &69.51\%  \\

$L^{feature}$=220                  & 69.16\%    &  69.35\%                        &72.13\%  \\

$L^{feature}$=225                  & 68.25\%    &  68.52\%                        &70.84\%  \\

$L^{feature}$=230                  & 68.65\%    &  69.76\%                       &69.36\%  \\

$L^{feature}$=235                  & 65.93\%    &  66.55\%                        &68.02\%  \\

$L^{feature}$=240                  & 67.84\%    &  69.12\%                        &67.91\%  \\

$L^{feature}$=245                  & 65.32\%    &  66.38\%                        &66.03\%  \\

$L^{feature}$=250                  & 66.02\%    &  66.37\%                        &69.02\%  \\

$L^{feature}$=255                  & 65.32\%    &  65.82\%                        &68.43\%  \\

$L^{feature}$=260                  & 65.32\%    &  66.03\%                        &67.80\%  \\

$L^{feature}$=265                  & 65.72\%    &  66.19\%                        &68.33\%  \\

$L^{feature}$=270                  & 65.72\%    &  66.75\%                        &66.99\%  \\

$L^{feature}$=275                  & 64.71\%    &  65.85\%                        &65.63\%  \\

$L^{feature}$=275                  & 64.71\%    &  65.85\%                        &65.63\%  \\

$L^{feature}$=280                  & 64.71\%    &  65.28\%                        &66.69\%  \\

$L^{feature}$=285                  & 66.33\%    &  67.89\%                        &65.66\%  \\

$L^{feature}$=290                  & 65.12\%    &  66.12\%                        &66.69\%  \\

$L^{feature}$=295                  & 62.89\%    &  64.23\%                        &63.50\%  \\

$L^{feature}$=300                  & 64.71\%    &  65.96\%                        &65.03\%  \\

$L^{feature}$=350                  & 62.89\%    &  64.23\%                        &63.50\%  \\

$L^{feature}$=400                  & 62.89\%    &  64.23\%                        &63.50\%  \\

$L^{feature}$=450                  & 62.89\%    &  64.23\%                        &63.50\%  \\

$L^{feature}$=500                  & 62.89\%    &  64.23\%                        &63.50\%  \\

\bottomrule

\end{tabular}
}
\caption{Full results of LRP4RAG$_\textrm{SVM}$ under different $L^{feature}$ on RAGTruth$_\textrm{Llama-7B}$.}\label{tab:resampling-appendix}

\end{table*}

\section{Proof of Softmax Taylor Decomposition } \label{appendix-taylor}

Given the softmax function and its derivatives:
\[
s_j(\mathbf{x}) = \frac{e^{x_j}}{\sum_i e^{x_i}},
\]
\[
\frac{\partial s_j}{\partial x_i} = 
\begin{cases}
  s_j(1-s_j) & \text{for } i=j \\
  -s_j s_i & \text{for } i \neq j
\end{cases}.
\]

Perform the first-order Taylor expansion at \(\mathbf{x}_0\):
\begin{align*}
s_j(\mathbf{x}) &\approx s_j(\mathbf{x}_0) + \sum_i \frac{\partial s_j}{\partial x_i} \bigg|_{\mathbf{x} = \mathbf{x}_0} (x_i - x_{0}) \\
& \approx s_j(\mathbf{x}_0) + s_j(\mathbf{x}_0)(1 - s_j(\mathbf{x}_0)) (x_j - x_{0}) + \\
& \sum_{i \neq j} -s_j(\mathbf{x}_0) s_i(\mathbf{x}_0) (x_i - x_{0}).
\end{align*}

Simplifying this, we have:
\begin{align*}
s_j(\mathbf{x}) &\approx s_j(\mathbf{x}_0) + s_j(\mathbf{x}_0)(1 - s_j(\mathbf{x}_0)) (x_j - x_{0}) - \\
& s_j(\mathbf{x}_0) ( \sum_{i} s_i(\mathbf{x}_0) (x_i - x_{0}) - s_j(\mathbf{x}_0) (x_j - x_{0})) \\
 & \approx s_j(\mathbf{x}_0) + s_j(\mathbf{x}_0)(1 - s_j(\mathbf{x}_0)) (x_j - x_{0}) - \\
& s_j(\mathbf{x}_0) \sum_{i} s_i(\mathbf{x}_0) (x_i - x_{0}) + s_j(\mathbf{x}_0)^2 (x_j - x_{0}) \\
 &\approx s_j(\mathbf{x}_0) + s_j(\mathbf{x}_0) (x_j - x_{0}) - s_j(\mathbf{x}_0)  \\
& \sum_{i} s_i(\mathbf{x}_0) (x_i - x_{0}) 
\end{align*}
Noting that \(\sum_{i} s_i(\mathbf{x}_0) = 1\), we get:
\begin{align*}
s_j(\mathbf{x}) & \approx s_j(\mathbf{x}_0) + s_j(\mathbf{x}_0) ( x_j - \sum_{i} s_i(\mathbf{x}_0) x_i ) \\
& = s_j(\mathbf{x}_0) \left( x_j - \sum_{i} s_i(\mathbf{x}_0) x_i \right) + \tilde{b}_j.
\end{align*}

The final form can be written as:
\begin{align*}
f_j(\mathbf{x}) &= s_j \left( x_j - \sum_i s_i x_i \right) + \tilde{b}_j.
\end{align*}

\end{document}